\documentclass[lettersize,journal]{IEEEtran}

\usepackage{amsmath,amsfonts}
\usepackage{algorithmic}
\usepackage{algorithm}
\usepackage{array}
\usepackage{textcomp}
\usepackage{stfloats}
\usepackage{url}
\usepackage{verbatim}
\usepackage{graphicx}
\usepackage{cite}

\usepackage{pgfplots}
\usepackage{subcaption}
\usepackage{array}
\usepackage{amsthm}
\usepackage{amssymb}
\usepackage{amsmath}
\usepackage[export]{adjustbox}
\usepackage{csvsimple}
\usepackage{makecell}
\usepackage{multirow}
\begin{document}

\title{Rethinking Generative Methods for Image Restoration in Physics-based Vision: A Theoretical Analysis from the Perspective of Information}

\author{Xudong KANG, Haoran Xie, Man-Leung Wong, Jing Qin}

\markboth{Journal of \LaTeX\ Class Files,~Vol.~14, No.~8, August~2021}%
{Shell \MakeLowercase{\textit{et al.}}: A Sample Article Using IEEEtran.cls for IEEE Journals}


\maketitle

\begin{abstract}
End-to-end generative methods are considered a more promising solution for image restoration in physics-based vision compared with the traditional deconstructive methods based on handcrafted composition models. However, existing generative methods still have plenty of room for improvement in quantitative performance. More crucially, these methods are considered black boxes due to weak interpretability and there is rarely a theory trying to explain their mechanism and learning process. In this study, we try to re-interpret these generative methods for image restoration tasks using information theory. Different from conventional understanding, we analyzed the information flow of these methods and identified three sources of information (extracted high-level information, retained low-level information, and external information that is absent from the source inputs) are involved and optimized respectively in generating the restoration results. We further derived their learning behaviors, optimization objectives, and the corresponding information boundaries by extending the information bottleneck principle. Based on this theoretic framework, we found that many existing generative methods tend to be direct applications of the general models designed for conventional generation tasks, which may suffer from problems including over-invested abstraction processes, inherent details loss, and vanishing gradients or imbalance in training. We analyzed these issues with both intuitive and theoretical explanations and proved them with empirical evidence respectively. Ultimately, we proposed general solutions or ideas to address the above issue and validated these approaches with performance boosts on six datasets of three different image restoration tasks.

\end{abstract}


\section{Introduction}
%Images captured by optical sensors or devices (like cameras) inevitably suffer from visual degradations caused by both internal (like noise, blur, aliasing, and compression artifact inside the camera) or external (such as rain, fog, haze, and other weather distortions) factors and can only reflect limited information of the observed scenes. Image restoration in physics-based vision (such as image denoising, dehazing, and deraining) has long been studied as a set of foundational but impactful tasks in computer vision, which attempts to remove these visual degradations and recover the captured scenes with clean backgrounds or of a higher visual quality.
%Images captured by cameras inevitably suffer from visual degradations caused by both internal (like noise, blur, aliasing, and compression artifact inside the camera) or external (such as rain, fog, haze, and other weather distortions) factors and can only reflect limited information of the observed scenes \cite{ImageRestoration}. Image restoration in physics-based vision (such as image denoising \cite{ImageDenoisingReview2,ImageDenoisingReview3}, dehazing \cite{ImageDehazingReview2}, and deraining \cite{ImageDerainingReview1}) has long been studied as a set of foundational tasks in computer vision, which attempts to remove these visual degradations and recover the captured scenes with clean backgrounds or of a higher visual quality.

\IEEEPARstart{I}{mages} captured by cameras inevitably suffer from visual degradations caused by both internal (like noise, blur, aliasing, and compression artifact inside the camera) or external (such as rain, fog, haze, and other weather distortions) factors and can only reflect limited information of the observed scenes \cite{ImageRestoration}. Image restoration in physics-based vision (such as image denoising \cite{ImageDenoisingReview2,ImageDenoisingReview3}, dehazing \cite{ImageDehazingReview2}, and deraining \cite{ImageDerainingReview1}) has long been studied as a set of foundational tasks in computer vision, which attempts to remove these visual degradations and recover the captured scenes with clean backgrounds or of a higher visual quality.

%Affected by complex physics systems, image restoration requires not only the simulation of the visual degradations (like noise, haze, and rain) but also the handling of how these degradations integrate with the background scenes to form the captured images. Recent advances in image restoration methods apply the representation learning idea of deep neural networks to simulate the complicated patterns of visual degradations without engineering features. However, regarding how these patterns integrate with the background, considerable methods tend to assume these degradations are just independent layer(s) that are linearly added onto the backgrounds or try to design various composition models to manually describe their integrations. These methods, noticeably, tend to be hypothetical models which are handcrafted based on human observation, statistical understanding, or prior knowledge under ideal conditions. They may not truly reflect the real-world scenarios, or may even involve human bias, contributing to the performance gap between the models evaluating on synthetic datasets and in actual practices.
With complex physics systems involved, image restoration requires not only the simulation of the visual degradations (like noise, haze, and rain) but also the handling of how these degradations integrate with the background scenes to form the captured images \cite{ImageDenoisingReview4,ImageDehazingReview3,ImageDerainingReview1}. Recent advances in image restoration methods apply the representation learning idea of deep neural networks to simulate the complicated patterns of visual degradations without engineering features. However, regarding how these patterns integrate with the background, considerable methods still tend to rely on handcrafted composition models that are manually designed to describe their integrations \cite{Dehaze_AtmosphericModel1,Dehaze_AtmosphericModel2,JORDER_Derain,DAFNet_Derain}. These methods, noticeably, tend to be hypothetical models which are handcrafted based on human observation, statistical understanding, or prior knowledge under ideal conditions. They may not truly reflect the real-world scenarios, or may even involve human bias, contributing to the performance gap between the models evaluating on synthetic datasets and in actual practices.

Generative methods are considered more promising solutions for image restoration tasks, which allow end-to-end simulation of the entire restoration processes using Deep Generative Models (DGMs) \cite{DGM1,DGM2,DGM3} without handcrafting composition models. Compared with the deconstructive idea above, generative methods also have better support in completing damaged / lost information, lighter-weight models, higher generalization-ability as well as many other advantages. Therefore, a growing number of recent studies start to apply generative methods to various image restoration tasks \cite{DGM_ImageRestoration1,GAN_ImageRestoration}.

%Nevertheless, existing GAN-based deraining methods tend to be direct applications of those general GAN models, leaving huge performance gaps compared with the handcrafted-model-based methods. Since these GANs models are not originally designed for single image deraining or relevant tasks, they are not supposed to be directly applicable and may even end up with convergence failure during the training processes.
Nevertheless, many existing generative methods still tend to have ample room to be improved in the quantitative performance compared with those deconstructive methods using handcrafted composition models, or may require more training data to achieve competitive performance with state-of-the-art results on many image restoration tasks.

Another problem lies in the interpretability of these generative methods: unlike deconstructive methods whose mechanisms are intuitively explainable, generative methods tend to be purely data-driven and can be a black box, where both the patterns of visual degradations as well as how they integrate with the backgrounds are learned inside the DGMs. There seems to be no solid theory specified for image restoration tasks that can explain the learning behavior inside these models nor understand its reliability.

%%%%%We found that: existing generative methods, especially GAN-based methods, tend to be simple applications of the DGMs designed for conventional generation tasks. Whereas, these general DGMs may not be directly suitable for the image restoration tasks and may remain several issues that result in the performance gaps above.

%%%%%Regarding the related theory, we also noticed that: conventional understanding tends to consider the generative methods in the image restoration tasks only as an information extraction process, where the network models simply attempt to optimize the extracted representation for better restoration (Fig. \ref{fig:conventional_understanding}). However, we consider in the actual scenario, a considerable proportion of fine-grained details may be retained intactly throughout the network without abstraction across layers, and there may exist a certain amount of background information in the target outputs that is absent / missed from the input images (Fig. \ref{fig:proposed_interpretation}). These two sources of information may also play essential roles in many restoration tasks and require learning / optimization correspondingly in the learning process.

%\iffalse
\begin{figure}[t]
    \centering
    \begin{subfigure}[b]{1.0\linewidth}
        \centering
        \begin{tikzpicture}[scale=0.5, every node/.style={scale=0.7}]
            \node[inner sep=0pt] (img_in) at (-7,0) {\includegraphics[width=2cm]{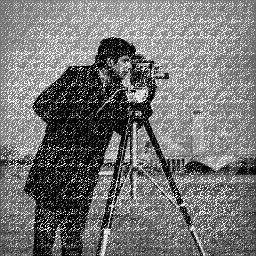}};
            \node[align=center, font=\scriptsize] at (-7, -2) {input image};
            \node at (-7, -2.8) {$X$};

            \node[inner sep=0pt, minimum width=2cm] (dnn) at (-2.4, 4) {};
            \fill[blue!5] ([yshift=12mm]dnn.north west)--([yshift=12mm]dnn.north east)--([yshift=-12mm]dnn.south east)--([yshift=-12mm]dnn.south west)--cycle;
            \node[inner sep=0pt, minimum width=2cm] at (-2.4, 4) {\includegraphics[width=1.4cm]{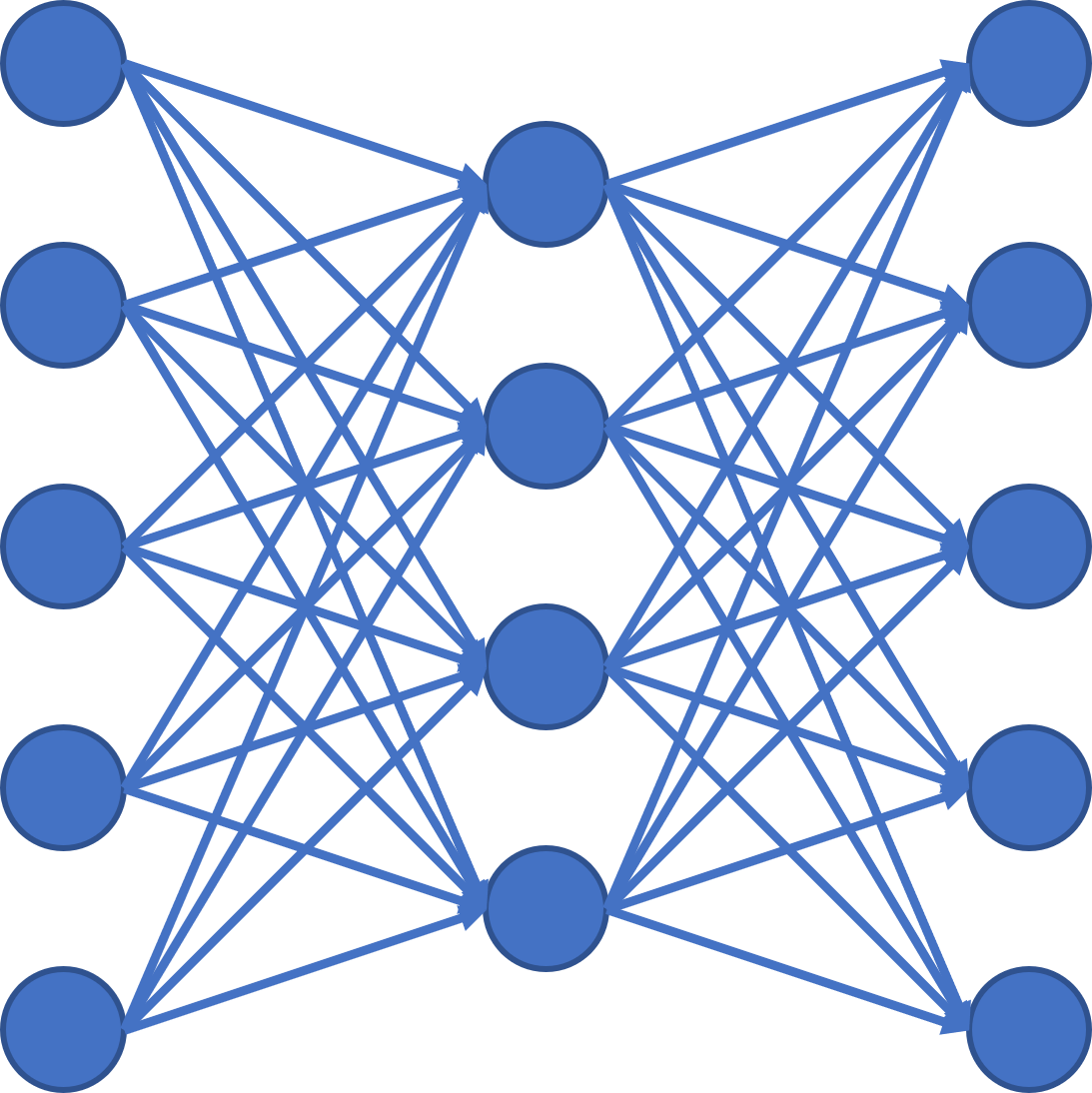}};
            \node[align=center, font=\scriptsize] at (-2.4, 2.3) {deep neural network};
            
            \node[inner sep=0pt, minimum width=2cm] (composition) at (-2.4, 0) {};
            \fill[gray!40] ([yshift=14mm]composition.north west)--([yshift=14mm]composition.north east)--([yshift=-14mm]composition.south east)--([yshift=-14mm]composition.south west)--cycle;
            \node[inner sep=0pt, minimum width=2cm] at (-2.4, 0) {\includegraphics[width=1.5cm]{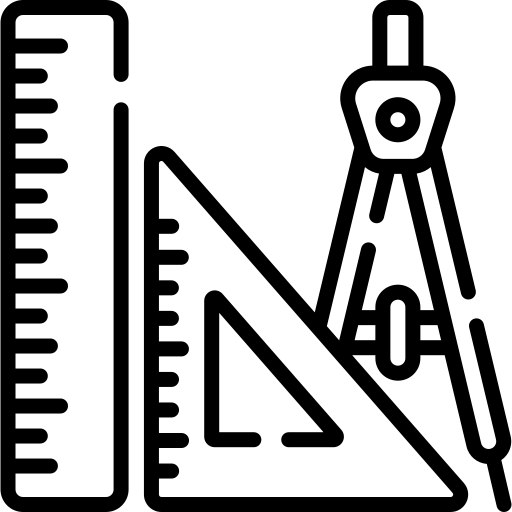}};
            \node[align=center, font=\scriptsize] (composition_label) at (-2.4, -2) {handcrafted \\[-0.6mm] composition models};

            \node[inner sep=0pt] (noise) at (2.3, 0){\includegraphics[width=2cm]{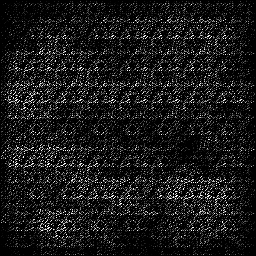}};
            \node[align=center, font=\scriptsize] at (2.3, -2) {extracted degradations};
            \node at (2, -2.8) {$X \mid \tilde{Y}$};

            \node[inner sep=0pt] (img_out) at (7,0) {\includegraphics[width=2cm]{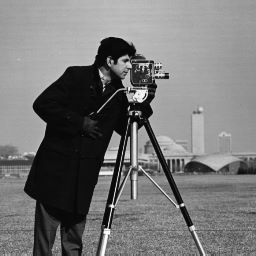}};
            \node[align=center, font=\scriptsize] at (7, -2) {restoration output};
            \node (img_out_label) at (7, -2.8) {$\tilde{Y}$};

            \draw[-stealth, line width=1] (img_in) -- (-7, 4) -- (dnn);
            \draw[-stealth, line width=1] (dnn) -- (2.3, 4) -- (noise);
            \draw[-stealth, line width=1] (img_in) -- (composition);
            \draw[-stealth, line width=1] (noise) -- (composition);
            \draw[-stealth, line width=1] (composition_label) -- (-2.4, -3.9) -- (7, -3.9) -- (img_out_label);

        \end{tikzpicture}
        \caption{deconstructive methods}
        \label{fig:deconstructive_methods}
    \end{subfigure}
    \begin{subfigure}[b]{1.0\linewidth}
        \vspace{4mm}
        \centering
        \begin{tikzpicture}[scale=0.5, every node/.style={scale=0.7}]
            \node[inner sep=0pt] (img_in) at (-7,0) {\includegraphics[width=2cm]{images/icons/img_in.png}};
            \node[align=center, font=\scriptsize] at (-7, -2) {input image};
            \node at (-7, -2.8) {$X$};

            \node[inner sep=1pt, minimum width=2cm] (encoder) at (-3, 0) {};
            \fill[green!10] ([yshift=12mm]encoder.north west)--([yshift=12mm]encoder.north east)--([yshift=-12mm]encoder.south east)--([yshift=-12mm]encoder.south west)--cycle;
            \node[inner sep=1pt, minimum width=2cm] at (-3, 0) {\includegraphics[width=1.5cm]{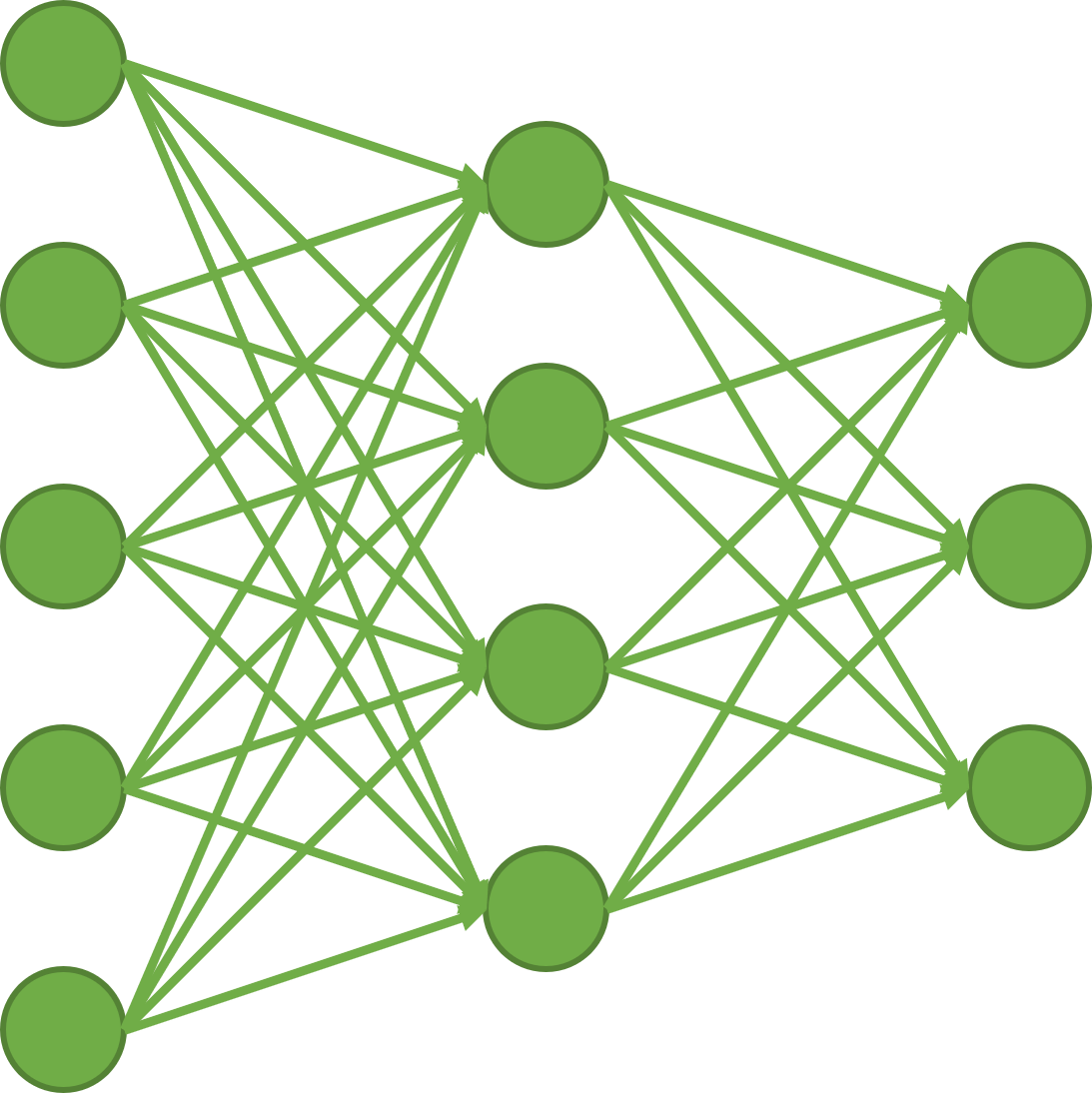}};
            \node[align=center, font=\scriptsize] at (-3, -2) {encoder network};

            \node[inner sep=0pt] (representation) at (0,0) {\includegraphics[width=0.65cm]{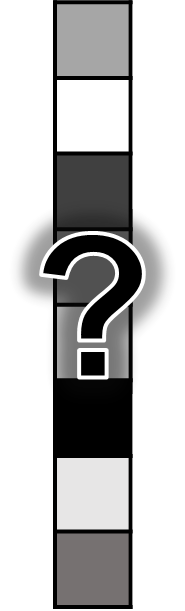}};
            \node[align=center, font=\scriptsize] at (0, -2) {latent space \\[-0.6mm] representation};
            \node at (0, -2.8) {$\tilde{X}$};

            \node[inner sep=1pt, minimum width=2cm] (decoder) at (3, 0) {};
            \fill[yellow!10] ([yshift=12mm]decoder.north west)--([yshift=12mm]decoder.north east)--([yshift=-12mm]decoder.south east)--([yshift=-12mm]decoder.south west)--cycle;
            \node[inner sep=1pt, minimum width=2cm] at (3, 0) {\includegraphics[width=1.5cm]{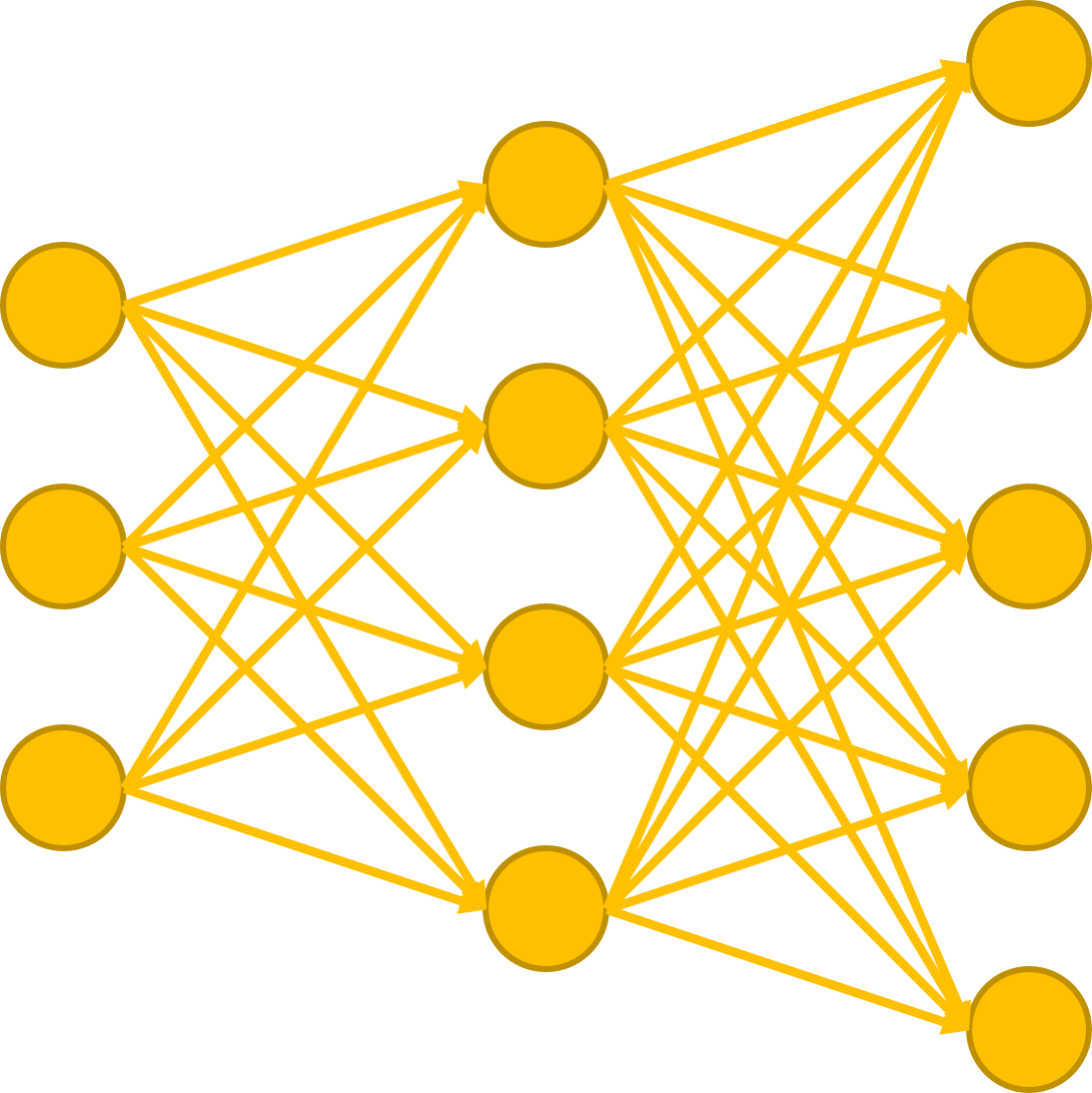}};
            \node[align=center, font=\scriptsize] at (3, -2) {decoder network};

            \node[inner sep=0pt] (img_out) at (7,0) {\includegraphics[width=2cm]{images/icons/img_out.png}};
            \node[align=center, font=\scriptsize] at (7, -2) {restoration output};
            \node at (7, -2.8) {$\tilde{Y}$};

            \draw[-stealth, line width=1] (img_in) -- (encoder);
            \draw[-stealth, line width=1] (encoder) -- (representation);
            \draw[-stealth, line width=1] (representation) -- (decoder);
            \draw[-stealth, line width=1] (decoder) -- (img_out);
        \end{tikzpicture}
        \caption{generative methods}
        \label{fig:generative_methods}
    \end{subfigure}
    \caption{Comparison between deconstructive methods and generative methods for image restoration}
    \label{fig:deconstructive_vs_generative}
\end{figure}
\begin{figure}[t]
    %\small
    \centering
    \begin{subfigure}[b]{1.0\linewidth}
        \centering
        \begin{tikzpicture}[scale=0.5]
            \centering
            \node at (-4.5, 0) {};
            \node at (15, 0) {};
            
            \node (Y) at (-3, 0) {$Y$};
            \node (X) at (-1, 0) {$X$};
            \node (X_tilde) at (1, 0) {$\tilde{X}$};
            \node (Y_tilde) at (3, 0) {$\tilde{Y}$};
            
            \draw[-stealth, line width=1] (Y) -- (X);
            \draw[-stealth, line width=1] (X) -- (X_tilde);
            \draw[-stealth, line width=1] (X_tilde) -- (Y_tilde);
            \node[anchor=west] at (5, 0) {
                $\displaystyle
                    \begin{cases}
                      {\min I(X; \tilde{X})} \\
                      \\
                      {\max I(Y; \tilde{X})} 
                    \end{cases}
                $
            };
        \end{tikzpicture}
        \vspace*{-5mm}
        \caption{Conventional Understanding}
        \label{fig:conventional_understanding}
    \end{subfigure}
    \vfill
    \begin{subfigure}[b]{1.0\linewidth}
        \centering
        \vspace*{3mm}
        \begin{tikzpicture}[scale=0.5]
            \node at (-4.5, 0) {};
            \node at (15, 0) {};
            
            \node (Y) at (-3, 0) {$Y$};
            \node (X) at (-1, 0) {$X$};
            \node (X_tilde) at (1, 1.8) {$\tilde{X}$};
            \node (Y_tilde) at (3, 0) {$\tilde{Y}$};
            
            \draw[-stealth, line width=1] (Y) -- (X);
            \draw[-stealth, line width=1] (X) -- (X_tilde);
            \draw[-stealth, line width=1] (X_tilde) -- (Y_tilde);
            \draw[-stealth, line width=1] (X) -- (Y_tilde);
            \draw[dashed, -stealth, red, line width=1] (Y) -- (-3, -1.5) -- (3, -1.5) -- (Y_tilde);
            \node[anchor=west] at (5, 0) {
                $\displaystyle
                    \begin{cases}
                      {\min I(X; \tilde{X}; \tilde{Y})} \\
                      {\max I(X \mid \tilde{X}; \tilde{Y})} \\
                      {\max I(Y \mid X; \tilde{Y})}
                    \end{cases}
                $
            };
        \end{tikzpicture}
        %\vspace*{0.1mm}
        \caption{Proposed Interpretation}
        \label{fig:proposed_interpretation}
    \end{subfigure}
    \caption{Information flows and optimization objectives of generative methods for image restoration tasks in conventional understanding versus our proposed interpretation}
    \label{fig:information_flow_and_objectives}
\end{figure}
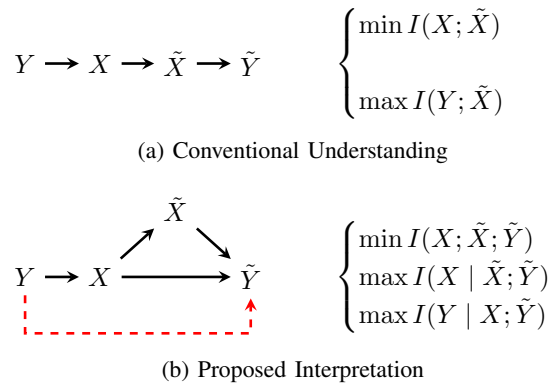

In this study, we noticed that: the conventional understanding tends to consider the generative methods in image restoration tasks only as an information extraction process, where the network models simply attempt to optimize the extracted representation for better restoration (Fig. \ref{fig:conventional_understanding}). However, we consider in the actual scenario, a considerable proportion of background information may be retained intactly throughout the network without abstraction across layers, and there may exist a certain amount of fine-grained details of the backgrounds in the target outputs that are absent / missed from the input images (Fig. \ref{fig:proposed_interpretation}). Based on this hypothesis, we analyzed the information flow in the generative methods of image restoration and affirm all three sources of information above are involved in generating the restoration results. By extending the information bottleneck principle, we re-interpret the learning process of DGMs in these generative methods: we deduced that the three sources of information above are to learn / optimized to approximate (i) the features / patterns of the visual degradations; (ii) background pixels / information to be retained in the restoration results; and (iii) fine-grained details or background information that is damaged / lost in the input images; respectively.
%(Section \ref{sec:information_analysis})

%%%%%Using this theoretical framework, we further analyzed the existing generative methods for image restoration and identified three major issues in the conventional DGMs when directly applied to the image restoration tasks, where: (i) these DGMs may contain over-invested abstraction processes; (ii) their network structures tend to inherently discard details information; and (iii) the training losses are actually optimizing two component objectives, which may contribute to gradient vanishing and imbalance of training in GAN-based models. We defined and formulate these issues with both intuitive and theoretical explanations (Section \ref{}). Then we provided empirical evidence and experiment results to prove their existence respectively, as well as to support and validate our theoretical framework (Section \ref{}).
Using this theoretical framework, we further found that: existing generative methods in image restoration tasks tend to be direct applications of DGMs designed for conventional generation tasks, where we identified three major issues in these conventional DGMs that may result in the performance gaps above: (i) these DGMs often contain over-invested abstraction processes; (ii) their network structures may inherently discard details information; and (iii) the loss functions for training tend to optimize two different component objectives, which may contribute to gradient vanishing and imbalance of training in GAN-based models. We analyzed and formulate these issues with both intuitive and theoretical explanations. Then we provided empirical evidence and experiment results to prove their existence respectively, as well as to support and validate our theoretical framework.

%(i) DGMs designed for conventional generation tasks involve over-invested abstraction processes than those required for learning the features / patterns of visual degradations in image restoration tasks; (ii) network structures of these models tend to inherently discard details information and may require specific design to support retaining intact information from the source inputs to the generated outputs; (iii) existing measures for evaluating generated results  We explain and formulate the above above issues both intuitively and theoretically. Then, we provided empirical evidence both quantitatively and qualitatively to prove the existences of these issues respectively as well as to to support and validate our theoretical framework.

%To address these issues, we propose general solutions or suggestions in Section \ref{section: solutions}, including optimization of network structure, enhancing details extraction, accumulation, and retention, as well as alternating the measures of the loss function. Corresponding methods are examined with experiments in Section \ref{section: experiment_general}.
Ultimately, we gave general solutions or ideas to address the above issues and to improve the performance of generative methods for image restoration, such as optimizing network structure, enhancing details extraction, accumulation, and retention, as well as using more sensitive measures of loss with pre-training. Then we validated these approaches with performance boosts on six datasets of different image restoration tasks, including image denoising, dehazing, deraining, and the hybrid of rain and haze removal.

To sum up, this study contributes in the following aspects:
\begin{itemize}
    %\item we elaborate the theory of information in the generative methods and analyze the flow and sources of information involved in generating the deraining results, which can be helpful for analyzing models of image deraining and relevant tasks;
    \item by revealing the sources and flow of information in these models, we elaborated the theory of generative methods in image restoration tasks and proposed an information-theoretic framework to explain the learning behaviors, optimization objectives, and their corresponding optimal information boundaries, which can be helpful for the analysis and design of relevant models;
    %\item we consider GAN models for general tasks may not be directly applicable to the tasks of image deraining, and we identified three major issues behind their performance gaps (i.e. over-invested abstraction process, structural discard of details, and, imbalance of training objectives) and proved with corresponding empirical evidence;
    \item we analyzed existing generative methods and identified three key issues in the direct application of conventional DGMs to image restoration tasks, where we provided intuitive analysis, theoretical explanations, and proofs with empirical evidence respectively;
    %\item we proposed solutions or suggestions to improve GAN models for the tasks of single image deraining, which can also be generalized to other image-to-image generation tasks.
    \item we proposed general solutions for the above issues, showed the ways to improve generative methods for image restoration tasks, and validated them on six datasets of three different image restoration tasks.
\end{itemize}

\section{Related Work}

%Based on the modeling of how rain layer(s) integrate with the background images, existing methods for single image deraining can be categorized into two groups: deconstructive methods try to describe the above integration by handcrafting composition models, while generative methods tend to learn this integration in a data-driven manner.

\subsubsection{Deconstructive Methods for Image Restoration}

Early studies of many image restoration tasks assume the visual degradations are linearly added onto the background scenes, and the related methods mainly focus on modeling these degradations for better removal by engineering their features \cite{ImageDenoisingReview2,ImageDenoisingReview3,ImageDecomposition_Derain,DiscriminativeSparseCoding_Derain,Bi-LayerOptimization_Derain,DirectionalGlobalSparseModel_Derain,GaussianMixtureModel_Derain}. Deep neural networks were later introduced to image restoration tasks and have now become the mainstream models for simulating these complicated patterns of visual degradations \cite{DeepCNN_Denoise,Dehazenet_Dehaze,JORDER_Derain,ImageDenoisingReview4,ImageDehazingReview3,ImageDerainingReview1}. Whereas, a growing number of recent studies started to figure out that these visual degradations may not be simply superposed onto the background scenes, and they proposed different theories and designed various composition models to describe how these degradations blend in with the backgrounds to form the captured images. Examples include the famous \textit{atmospheric scattering model} \cite{Dehaze_AtmosphericModel1,Dehaze_AtmosphericModel2} in the image dehazing, as well as the \textit{heavy rain model} \cite{JORDER_Derain} and the \textit{depth-aware rain model} \cite{DAFNet_Derain} in the image deraining task. However, all these methods still consider the visual degradations as independent layer(s) of pixels and try to manually deconstruct / interpret their integrations using human assumptions or statistical understandings based on limited data, which may involve human bias and fail to truly reflect the real-world situations.

\subsubsection{Generative Methods for Image Restoration}
Recent studies try to use DGMs to directly learn / simulate the end-to-end mappings of image restoration tasks without the need to understand their compositions or detailed mechanisms, which shows considerable advantages (summarized in Appendix A) compared with the deconstructive methods above. 

%%%%%The simplest form of generative methods is based on AEs, where a generator network is trained mainly by maximizing pixel-level similarities between restoration output and the target ground truth. Relevant methods have been applied to image denoising \cite{BlindAE_Denoise,AE_blur_Denoising}, deraining \cite{conditional_VAE_Derain_1,conditional_VAE_Derain_2,LPNet_Derain}, dehazing \cite{CAE_Dehaze,LCANet_Dehaze}, as well as other image restoration tasks \cite{AE_ImageRestoration_1,AE_ImageRestoration_2,AE_ImageRestoration_3}. However, it can be difficult for AE-based generative methods to learn high-level semantics knowledge for generating plausible results or may require extra domain-specific knowledge \cite{LPNet_Derain}.
As the simplest form of these generative methods, Autoencoders (AEs) \cite{AutoEncoder,SparseAutoEncoder,VAE} have been applied to image denoising \cite{BlindAE_Denoise,AE_blur_Denoising}, deraining \cite{conditional_VAE_Derain_1,conditional_VAE_Derain_2}, dehazing \cite{CAE_Dehaze,LCANet_Dehaze}, and other image restoration tasks \cite{AE_ImageRestoration_1,AE_ImageRestoration_2,AE_ImageRestoration_3}. However, it can be difficult for AE-based generative methods to learn high-level semantics knowledge for generating high fidelity results or may require extra domain-specific knowledge \cite{LPNet_Derain}.

Generative methods based on Generative Adversarial Networks (GANs) \cite{GAN,DCGAN,CGAN} can be regarded as an improved version of the above AE-based generative methods, which introduce an extra discriminator network with an adversarial training strategy to allow generating more eidetic results. In fact, most GAN models for image-to-image translation (like pixel2pixel \cite{pixel2pixel} and CycleGAN \cite{cycleGAN}) can be directly adopted in many image restoration tasks and can obtain plausible results, but their quantitative performance on the benchmarks tend to be less satisfactory. Many existing GAN-based image restoration methods attempt to reduce these performance gaps by modifying these basic architectures \cite{GAN_Denoise,BlindDenoiseGAN,UnpairedDenoiseGAN,ID-CGAN_Derain,AttentiveGAN_Derain,CGAN_Dehaze,DHSGAN_Dehaze,CycleDehaze}. However, we consider many of them only made minor changes where key issues in these conventional GANs seem to be ignored or left unsolved. Some others only apply GANs as supplementary, where their network structures still tend to be based on the deconstructive idea or do not use end-to-end training \cite{HeavyRainRestorer_Derain,AODNet_Dehaze}.

Despite their promise in image restoration, many existing generative methods still tend to be direct applications of general DGMs for conventional generation problems, whose performances may be less competitive with the deconstructive methods or may require extra training data to converge.

\subsubsection{Conventional Interpretation of the Generative Methods}

%%%%%Deep neural networks have long been viewed as black boxes due to lacking interpret-ability \cite{DeepLearning,ExplainBlackBoxDeepLearningReview}. Especially for DGMs, the generation processes are fully conducted by the parameters inside the generator networks, where the internal operations and the reliability of the generated results are still unclear. Compared with the deconstructive methods that have been well-studied and whose mechanism can be easily and intuitively explained, generative methods in image restoration tend to be under-exploration. And, so far as we known, there is no solid theory to explain the mechanism or learning process inside the DGMs of these generative methods specified for image restoration tasks.
Unlike deconstructive methods, whose mechanisms can be easily explained, generative methods of image restorations have long been viewed as black boxes due to lacking interpretability. So far as we know, there is no solid theory to explain the mechanism or learning behaviors inside the DGMs of these generative methods for image restoration tasks.

%%%%%Whereas, some relevant studies try to intuitively interpret their learning processes. The most common understanding regards the generative methods of image restoration as a background extraction process, where the encoder part of the DGMs attempts to extract latent space representations / embeddings from the visually-degraded inputs that contain as much information about the target outputs (clean background scenes) as possible, and the decoder network simply reconstruct the restoration results using only the information from these representations. These studies deem that the restoration performance is fully determined by the quality of the extracted representation, where the higher quality of the representation extracted, the better restoration results can be obtained \cite{ERLNet_Derain}. \citeauthor{DRLENet_Derain} (\citeyear{DRLENet_Derain}) further extended this idea and disentangled the latent representations into task-relevant (background, contents \& details) and task-irrelevant (visual degradations) parts, where they interpret the learning process as to isolate the task-irrelevant factors so as to reduce the ambiguity of these learned representations. These interpretations consider the image restoration performance are fully determined by the encoder's capability in extraction background information from the source inputs and tend to ignore the training / optimization in the decoder network. 

Whereas, some related studies try to intuitively interpret their learning processes. The most common understanding regards the generative methods of image restoration simply as a background extraction process, which believes the restoration performance is fully determined by the quality of latent space representations / embeddings extracted by the encoders of DGMs \cite{ERLNet_Derain}. \cite{DRLENet_Derain} further extended this idea and disentangled the latent representations into task-relevant (background / contents to be restored) and task-irrelevant (visual degradations) factors, where they interpret the learning process as to isolate the task-irrelevant part so as to reduce the ambiguity of these learned representations. These conventional understandings deem the image restoration performance is fully dependent on the ``background-extract-ability" of the encoder networks, where the extracted representations only need to contain as much information about the target's contents / background scenes as possible, but they tend to ignore the training / optimization in the decoder networks.

\cite{Disentangled_Unsupervised_Deblur} also considers that the extracted representations consist of the above two kinds of features. But differently, they use two separate encoders to respectively learn each kind of feature, and, rather than suppressing the task-irrelevant information (visual degradations) before sending it to the decoder, they let the decoder trade-off between the two sources of representations. This interpretation steps closer to our findings. However, it does not take into account the differences between the two types of features in the levels of abstraction and the amount of required information, and it still considers that the source inputs contain all the information for restoring the target outputs, which we found to be less accurate.

Generally speaking, all the ideas above seems to be simple interpretations referred from the conventional understanding of DGMs for general domain transfer generation problem and may not accurately reflect the actual mechanism of generative methods in image restoration tasks.

\section{Information-theoretic Framework}
\label{sec:information_analysis}

Information Bottleneck (IB) Principle \cite{Information_Bottleneck_Deep_Learning_1,Information_Bottleneck_Deep_Learning_2} theoretically interprets the learning behaviors of general deep neural network models by employing an information-theoretic method. This theory explains the information flow and quantifies the optimization process with information.

In the conventional understandings, generative methods for image restoration are interpreted as an information extraction process about background contents from the source inputs. This can be directly explained using the IB theory: given a visually-degraded image $X$ to be fed into a DGM, its desired output $Y$ is regarded as the image of its corresponding clean background, which, in reverse, determines the basic information of $X$. Suppose we consider the network layers in the DGM as a whole, hence we define the representation obtained from the latent layers as $\tilde{X}$ and the final output from the DGM $\tilde{Y}$ as the estimated restored image in approximation to $Y$. Their dependency relationship can form a Markov Chain: $Y \to X \to \tilde{X} \to \tilde{Y}$ (same as Fig. \ref{fig:conventional_understanding}), where the optimization goal of the learning process is to maximize the mutual information between the extracted latent representation $\tilde{X}$ and the ideal output $Y$ while minimizing the mutual information between $\tilde{X}$ and the input $X$:

\begin{equation}
    \min [ I(X;\tilde{X}) - \beta I(Y;\tilde{X}) ]
\end{equation}

\noindent where $\beta$ is a positive Lagrange multiplier that trades-off between the two terms.

According to the Data Processing Inequality (DPI) \cite{DPI}, we can have the optimal information boundaries of this conventional interpretation of the learning process:

\begin{equation}
    I(Y;X) \geqslant I(Y;\tilde{X}) \geqslant I(Y;\tilde{Y})
\end{equation}

%\noindent where the first equality is satisfied if and only if $\tilde{X}$ is a sufficient statistic of $X$, which requires the encoder network to be powerful enough to extract all the mutual information $I(Y; X)$ in $\tilde{X}$, which is believed to be the total information about the contents (background scenes) from the inputs $X$ in the high-level latent space representation ; and, similarly, the second equality is satisfied if and only if $\tilde{Y}$ is a sufficient statistic of $\tilde{X}$, requesting the decoder network to pass the entire information it received to the outputs $\tilde{Y}$. In this way, the targeted mutual information $I(Y;\tilde{Y})$ can be maximized to reach $I(Y; X)$, which is believed to be the same as $H(Y)$. 
\noindent where the first equality is satisfied if and only if $\tilde{X}$ is a sufficient statistic for $Y$ based on $X$, which requires the encoder network to be powerful enough to fully extract the mutual information $I(Y; X)$ in its high-level embedding / latent space representation $\tilde{X}$, and, similarly, the second equality is satisfied if the decoder can pass the entire information it received to the output $\tilde{Y}$. In this way, the mutual information in the restoration result $I(Y;\tilde{Y})$ can be maximized to reach $I(Y; X)$, which, in these conventional understandings, is believed to contain all information of the target $Y$.

%However, we noticed that: the information about contents / background scenes in the inputs is supposed to be simply retained without the need for an abstraction process across the encoders' network layers. Moreover, in the actual practices, many commonly-used generator networks (such as U-Net \cite{UNet}) even provide structures like skip connections to allow passing this low-level information directly and intactly to the decoders for generation. Thus in this study, we consider that: besides the high-level features learned by the encoders $\tilde{X}$, considerable low-level information from the inputs $X$, such as the background pixels may probably be retained throughout the generator network.
However, we noticed that: the information about contents / background scenes in the inputs is supposed to be simply retained without the need for an abstraction process across network layers. Moreover, in actual practices, many commonly-used DGMs (such as U-Net \cite{UNet}) even provide structures like skip connections to allow passing this low-level information directly and intactly to the decoders / generation part of the models without going through the encoders. Thus in this study, we consider that: besides the high-level features learned by the encoders $\tilde{X}$, considerable low-level information from the inputs, such as the background pixels, may probably be retained throughout the networks of DGMs ($I(X \mid \tilde{X}; \tilde{Y})$).

%In addition, the conventional interpretation above assume the source inputs $X$ contain all the information required for restoring the targets $Y$ (i.e. $I(Y;X) == H(Y)$). But we deem that may not be the case in the real scenarios of many image restoration tasks: due to serious distortion like blurring and covering, some pixels of background in the observed image may not be recovered by using only the information from this single input. In fact, most data-driven generative models tend to more or less ``imagine" the missing contents based on external knowledge learned from other inputs or predictions calculated by the models' parameters \cite{Denoising_AE_generative,GAN}.

In addition, the conventional interpretation above assume the entirety of restoration targets $Y$ can be retrieved from their corresponding source inputs $X$. But in the real scenarios of many image restoration tasks, $X$ may not contain all the information required for restoring the targets $Y$ ($I(Y; X) \neq H(Y)$): some background pixels observed may be seriously distorted, blurred, damaged or may even be completely covered by the visual degradations. Thus, relevant information may have already been lost and may not be recovered using only the information from a single input. In fact, most data-driven generative models tend to more or less ``imagine" the missing contents based on the predictions of network parameters or external knowledge learned from multiple inputs \cite{Denoising_AE_generative,GAN}.

%Moreover, besides the high-level features learned by the encoder $\tilde{X}$, some low-level information from the inputs $X$, such as the background pixels may probably be retained throughout the generator network. To better reconstruct the fine-grained details in the generated images, some generator networks even provide structures like skip connections to assist in passing this intact information directly to the generating part of the network.

%other than the learned features about the "noise", such as the sufficient statistics describing the distribution of the rain patterns, to reconstruct the fine-grained details in the generated images,

Therefore, to sum up, we consider three sources of information are involved in generating the restoration results $\tilde{Y}$:

%(i) extracted high-level information, (ii) retained low-level information, and (iii) external information that is absent from the source inputs

\iffalse

\begin{enumerate}
    \item high-level information that is provided by the extracted features / latent space representations $\tilde{X}$ from the encoder network / feature extraction model: $I(\tilde{X}; \tilde{Y})$;
    \item low-level information in source inputs $X$ that pass directly through the skip connections or retained intactly in the results $\tilde{Y}$: $I(X \mid \tilde{X}; \tilde{Y})$;
    \item external information that is absent from the source inputs but later involved by the parameters of networks in the restoration results: $H(\tilde{Y} \mid X, \tilde{X})$.
\end{enumerate}

\fi

\begin{enumerate}
    \item high-level information from the feature embeddings / latent space representations $\tilde{X}$ that are extracted by the encoder networks / feature extraction models: $I(\tilde{X}; \tilde{Y})$;
    \item low-level information in the source inputs $X$ that pass directly through the skip connections or are retained intactly in the results $\tilde{Y}$: $I(X \mid \tilde{X}; \tilde{Y})$;
    %\item external information that is absent from the source inputs but later involved by the parameters of networks in the restoration results: $H(\tilde{Y} \mid X, \tilde{X})$.
    \item external information involved by the parameters of networks in the restoration results without coming from the source inputs: $H(\tilde{Y} \mid X, \tilde{X})$.
\end{enumerate}

%Based on the above insights, we re-interpreted the flow of information as Figure \ref{fig:proposed_interpretation}, where the Markov Chain in the conventional understanding above no longer stands, and we deduced that the learning processes are to optimize the above three sources of information correspondingly (See Appendix B for more detailed analysis and explanations). By analyzing the possible ranges of each part of the information, we can derive the overall training objectives and the corresponding optimization boundaries for each of its components as follow (derivation and proof are attached in Appendix C):

Based on the above insights, we re-interpreted the flow of information as Figure \ref{fig:proposed_interpretation}, where we deduced that the learning processes are to optimize the above three sources of information correspondingly (See Appendix B for more detailed analysis and explanations). By analyzing the possible ranges of each part of the information, we can derive the overall training objectives and the corresponding optimization boundaries for each of its components as follow (derivation and proof are attached in Appendix C):

\begin{equation}
    \min [ I(X; \tilde{X}; \tilde{Y}) - \beta_{1} I(X \mid \tilde{X}; \tilde{Y}) - \beta_{2} I(Y \mid X; \tilde{Y}) ]
\end{equation}

\noindent s.t.

\begin{equation}
    \begin{cases}
      I(X; \tilde{X}; \tilde{Y}) \geqslant -H(X \mid Y) \\
      I(X \mid \tilde{X}; \tilde{Y}) \leqslant H(X) \\
      I(Y \mid X; \tilde{Y}) \leqslant H(Y \mid X)
    \end{cases} 
\end{equation}

\noindent where $\beta_{1}$ and $\beta_{2}$ are positive coefficients.

In simple terms, we interpret the internal process and learning behaviors of the generative methods in image restoration as follows:

\iffalse
\begin{enumerate}
    \item we disentangled the information passed from the sources inputs to the restoration results into different abstraction levels: with sufficient amounts of information allowed to pass through the network, the high-level information $I(\tilde{X}; \tilde{Y})$ in the latent representation extracted by the encoders, will be optimized to approximate the features / patterns of visual degradations ($-H(X \mid Y)$), while the removal of this information happens in the generation process of the decoder network (rather than in the extraction process of the encoder network);
    \item the contents / background scenes information to be restored is considered low-level information that can be retained throughout the network without going through the abstraction process across the encoder network, and this part of the information $I(X \mid \tilde{X}; \tilde{Y})$ will be optimized to approach the intact source inputs $H(X)$;
    %\item besides the information extracted / retained from the source inputs, the decoder network may also learn and involve external knowledge to approach / complete the information relevant to the targets but is absent in the source inputs $H(Y \mid X)$.
    \item besides the information extracted or retained from the inputs, the decoder network may also involve external knowledge in its restoration outputs $H(\tilde{Y} \mid X, \tilde{X})$, which is optimized to approach / complete the information of the targets but is absent in the inputs $H(Y \mid X)$.
\end{enumerate}
\fi

\begin{enumerate}
    \item rather than doing only the background extraction, the encoder networks process and deliver both the features / patterns of visual degradations $H(X \mid Y)$ and the information of contents / background scenes $I(X \mid \tilde{X}; \tilde{Y})$ in the sources images if sufficient amounts of information are allowed to pass, while the removal of $H(X \mid Y)$ happens in the generation process of the decoder networks (rather than in the extraction process of the encoder networks);
    \item in the encoder parts of networks, the two kinds of information above can be disentangled according to their differences in levels of abstraction, and therefore are processed by different structures of the networks: the high-level information $I(\tilde{X}; \tilde{Y})$ in the latent representation extracted by the encoders, will be optimized to approximate the visual degradations ($H(X \mid Y)$), while the contents / background scenes information to be restored is considered low-level information that can be retained throughout the network without going through the abstraction process across the encoder networks, and this part of the information $I(X \mid \tilde{X}; \tilde{Y})$ will be optimized when the intact amount of information of the source input $H(X)$ can be passed;
    \item besides the information extracted or retained from the inputs, the decoder network may also involve external knowledge in its restoration outputs $H(\tilde{Y} \mid X, \tilde{X})$, which is optimized to approach / complete the information of the targets but is absent in the inputs $H(Y \mid X)$.
\end{enumerate}

%suppose the decoder network is powerful enough to pass all information to the generated output $\tilde{Y}$, then the first term of the above objective requires a powerful enough encoder to extract irrelevant information about $Y$ from $X$, the second term makes demands on the passing of information from $X$ directly to $\tilde{Y}$, while the last term request the decoder to learn information that can contain information from $H(Y \mid X)$. Theoretically, the maximum information that the first two terms together can reach is $I(Y; X)$, and the ideal learning target for the third term is to approximate $H(Y \mid X)$.

%Ideally, we consider the first term is to approximate $-H(X \mid Y)$, and the second term is to approximate $H(X)$.

\section{Existing Problems \& Analysis}
\label{sec:problem_definitions}

Many existing generative methods for image restoration tasks tend to be simple applications of general DGMs that was originally designed for conventional generation problems. According to the above theory, we can identify three critical issues (corresponds to the optimization of three information sources above) in the conventional DGMs that probably contribute to the performance gaps.

\subsection{Problem 1: Over-invested Abstraction Process}
\label{section: excessive_abstraction}

\noindent \textbf{Description:} \textit{Features / patterns of visual degradations in an image restoration task only require a specific level of abstraction for extraction / simulation and occupy only a certain amount of information. However, conventional DGMs tend to contain excessive abstraction processes, which may not help the performance of image restoration tasks, bring in unnecessary network parameters, and may even involve noises / irrelevant information.}

%Patterns of rain in the image deraining task are locally distributed and relatively lower-level features, while generators designed for conventional GANs problems involve excessive abstraction processes for learning global and higher-level features, which may not help with learning rain features when applied to the task of image deraining, contributing to excessive network parameters, irrelevant information involved, and even drops in deraining performances.

\noindent \textbf{Intuition / Observation:} DGMs designed for conventional generation problems are supposed to learn higher-level semantic features that globally span large pixel areas, while visual degradations in image restoration tasks tend to be locally distributed and relatively lower-level features according to Marr's definition \cite{Marr_Vision}.

\noindent \textbf{Analysis / Theoretical Explanation:} See Appendix D.

\subsection{Problem 2: Inherent Details Loss}
\label{section: inherent_details_loss}

\noindent \textbf{Description:} \textit{The network structures of the conventional DGMs do not support retaining intact inputs in the generated results, where low-level information may be discarded inherently in both extraction and generation processes. In image restoration tasks, this mainly corresponds to the loss of background information and fine-grained details, contributing to severe distortion and poor quantitative performance in the restoration results.}

\noindent \textbf{Intuition / Observation:} Traditional generation problems pay more attention to the high-level consistency of the generated results and encourage variations in the low-level details, but this can be fatal to the image restoration tasks.

\noindent \textbf{Analysis / Theoretical Explanation:} See Appendix E.

\subsection{Problem 3: Vanishing Gradient \& Imbalance Training}
\label{section: vanishing_gradient}

%\noindent \textbf{Description:} \textit{Unlike conventional generation problems, the same loss functions applied to image restoration tasks tend to optimize two uneven component objectives, thus may not provide smooth gradients for the continuous convergence of models and may encounter abrupt changes during the training process, contributing to vanishing gradient or even leading to an imbalanced updating between the generators and the discriminators in GAN-based generative methods.}
\noindent \textbf{Description:} \textit{Loss functions used in conventional generation problems tend to optimize two uneven component objectives when applied to image restoration tasks. Thus, they may no longer provide smooth gradients for the continuous convergence of models and may drop abruptly during the training process, contributing to vanishing gradient or even leading to an imbalance in the updating between the generators and the discriminators in GAN-based methods.}

%\noindent \textbf{Intuition / Observation:} In conventional generation tasks, the inputs are often independent of the target outputs (such as random noise) or do not contain much information about the targets. But for image restoration tasks, the source inputs and the target outputs share considerable similarities (like the vast majority of the same pixels of the backgrounds). Therefore, the models may converge much easier by utilizing this similar information in the inputs but may become difficult to learn the knowledge about the targets that are not involved in the inputs.
\noindent \textbf{Intuition / Observation:} In conventional generation tasks, the inputs are often independent of the target outputs (like random noise) or do not contain much information about the targets. But for image restoration tasks, the source inputs and the targets tend to share considerable similarities (like the majority of the same pixels of the backgrounds). Therefore, the models may converge much easier by utilizing this similar information but may become difficult to learn knowledge about the targets that are not involved in the inputs.

\noindent \textbf{Analysis / Theoretical Explanation:} See Appendix F.

\section{Solutions \& Methods}
\label{sec:solutions}

To improve the performance of generative methods for image restoration tasks, in this section, we indicate the general solutions / suggestions for the above problems as well as specific methods to validate them respectively.

\iffalse

In a bid to enhance the optimization of the three sources of information as well as to improve the performance of generative methods for image restoration tasks, we need to address the above three 

Here we proposed general solution to address the above issues, as well as specific methods to validate the above problems.

\begin{equation}
    \begin{cases}
      I(\tilde{X}; \tilde{Y}) & \to H(X \mid Y)\\
      I(X \mid \tilde{X}; \tilde{Y}) & \to H(X)\\
      H(\tilde{Y} \mid X, \tilde{X}) & \to H(Y \mid X)
    \end{cases}       
\end{equation}

\noindent where the first case pursuits the encoder network to be powerful enough for extracting all information from $X$ that is irrelevant to $Y$, which is the rain patterns, the second case requires intact information in $X$ can be passed to $\tilde{Y}$, and the last case makes demands on the decoder to learn extra information so as to complete the missing information of $Y$ in $X$ as plausible as it can (See appendix for more detailed explanations).

\fi

%%%%%To prevent the over-invested abstraction process, a shallow network with fully-convolutional down-sampling and skip connections can be adopted as the backbone generator for the deraining task. Here we consider a shallow U-Net \cite{UNet,pixel2pixel} with $N_{saturated}$ layers of down-and-up-sampling is sufficient to extract most of the high-level information required.
To prevent the over-invested abstraction process, we need to investigate the minimum requirements for extracting / simulating the corresponding visual degradations in the image restoration tasks, and therefore remove the unnecessary abstraction process and redundant network parameters. For DGMs based on Convolutional Neural Networks (CNN), this process of abstraction is often realized by the down-and-up-sampling mechanism. Thus, we consider for each kind of visual degradation, there exists a specific number of down-and-up-sampling layers and a certain dimensionality of the latent representations that can be sufficient to fully simulate / extract the patterns / features of this degradation, where the layers or dimensions larger than these numbers may do no good to the restoration performances.

%%%%%To reduce the inherent details loss, we need to handle the discard of low-level information both before the decoder ($I(X \mid \tilde{X}; M) \ll H(X \mid \tilde{X})$), as well as inside the decoder networks ($I(M; \tilde{Y}) \ll H(M)$). However, we cannot directly increase the amount of low-level information that passed to the decoder $I(X \mid \tilde{X}; M)$ by simply modifying the skip connections without affecting $H(\tilde{X})$. As an alternative, we proposed to increase the total amount of information in the inputs to achieve this goal. According to $H(X \mid \tilde{X}) = H(X) - I(X; \tilde{X})$, where $I(X; \tilde{X})$ we consider is upper-bounded by the amount of information of the rain features, which can be constant due to excessive $H(\tilde{X})$, thus, increasing $H(X)$ may help to increase $H(X \mid \tilde{X})$ (see appendix for more detailed explanations).
%%%%%To alleviate details loss that happens inside the decoder, we also proposed to enhance the decoder network to better retain information in their generated outputs. Referring to the generative methods of super-resolution for generating higher quality images, we modified the up-sampling process of existing decoder networks by adopting the sub-pixel convolution \cite{Sub-pixel_Conv} (SCU) as the enhancement (see appendix for more detailed explanations).

To reduce the inherent details loss, we need to handle the discard of low-level information both before and inside the decoder networks. For the first parts of information loss, a global skip-connection that can pass intact inputs directly to the decoder networks may solve. But in CNN-based DGMs, this may not be easily applicable without affecting the latent representations $H(\tilde{X})$. As a more general solution, we proposed increasing the total amount of information in the inputs so as to guarantee that more information can be retained. According to $H(X \mid \tilde{X}) = H(X) - I(X; \tilde{X})$, where $I(X; \tilde{X})$ can be regarded as a constant (upper-bounded by the amount of information of the visual degradations $H(X \mid Y)$), increasing $H(X)$ may help to improve $H(X \mid \tilde{X})$. More specifically, to achieve this goal, we put an information accumulation (\textit{InfoAccum}) module before the DGMs, which enhances the extraction and accumulation of the inputs' information before sending it to the encoder, and the number of layers in this module can reflect the total amount of this accumulated information (see Appendix G for more details and relevant discussion). As for the second part of details loss that happens inside the decoder network, we need to search for a decoder network that can be powerful enough to: (i) retain all information it received in its outputs, (ii) parse the latent representation extracted by the encoder and remove the information of visual degradations, as well as (iii) to learn external knowledge for completing the missing details. For CNN-based DGMs, we consider an enhancement in the upsampling methods of the decoder network may help.

%%%%%On account of the vanishing gradient and imbalance during the training process, we proposed to use the LSGAN loss \cite{LSGAN} to replace the traditional GAN loss based on JS-divergence \cite{GAN}. It can reduce the problem of vanishing gradient when the distributions between the targets the generated results are fairly close to each other, allowing further convergence of GAN models.
As for the vanishing gradient and imbalance when training GAN models, we suggest using more sensitive measures of loss functions in the later stages of training and consider pre-training on image reconstruction (for autoencoders or generators in GANs) and on extra datasets (for discriminators in GANs) may help to accelerate convergence and balanced two models in a GAN architecture.

\section{Experiments}
\label{sec:experiments}

%Here we provided empirical evidence to prove the above three problems respectively and validate our proposed solution as well as our interpretation (theoretical framework) through general experiments on six benchmarks of different image restoration tasks.
Here we provided empirical evidence to prove the above three problems respectively and validated our proposed solutions as well as the theoretical framework with general experiments on six benchmarks of different image restoration tasks.

\subsubsection{\textbf{Empirical Evidence of Problem 1}}
\label{subsec:evidence_1}

To prove the existence of over-invested abstraction processes, we investigate the image restoration performances of DGMs with different levels of abstraction. Here we adopted two common types of backbone DGMs for image-to-image translation: convolutional encoder-decoders without skip connection \cite{ConvEnDecoder_segmentation} (En/Decoder), and U-Net \cite{UNet} (UNet), each of them with different numbers of down-and-up-sampling layers respectively. We train and test these methods on the benchmarks of three different image restoration tasks (see Appendix G for more implementation details). Experiment results (Fig. \ref{fig:excessive_abstraction}) verified that: the number of down-and-up-sampling layers $N$ tends to be saturated at certain values $N_{saturated}$, where continuing increasing $N > N_{saturated}$ does not improve the performance of the model (UNet), or may even cause a performance drop (En/Decoder). Even more interesting is that more complicated visual degradations seem to require higher levels of abstraction process (Tab. \ref{tab:saturated_layers_on_different_tasks}), which intuitively makes sense.

\begin{figure}[!htbp]
    \centering
    \begin{adjustbox}{center}
        \begin{tikzpicture}
            \begin{axis}
                [
                width=0.8\linewidth,
                height=2in,
                xmin=0.5, xmax=8.5, xtick={1,...,8},
                xlabel={\scriptsize number of down-and-up-sampling layers},
                ylabel={\scriptsize de-raining performance (SSIM)},
                xlabel near ticks,
                ylabel near ticks,
                xlabel style = {inner sep=0pt},
                ylabel style = {inner sep=0pt},
                ticklabel style = {font=\tiny},
                legend style={at={(0.02,0.02)}, anchor=south west, font=\tiny}
                ]
                \addplot table [x=num_layers, y=unet-5, col sep=comma] {data/excessive_abstraction.csv};
                \addplot table [x=num_layers, y=conv_encoder_decoder, col sep=comma] {data/excessive_abstraction.csv};
                \legend{UNet, En/Decoder}
            \end{axis}
        \end{tikzpicture}
    \end{adjustbox}
    \captionof{figure}{Image restoration performances with different numbers of down-and-up-sampling layers. Here we demonstrates the deraining performance on Rain800 datasets, while similar patterns are observed on other datasets for different image restoration tasks.}
    \label{fig:excessive_abstraction}

    \vspace{0.2in}

    \renewcommand{\arraystretch}{1.5}
    \small
    \begin{adjustbox}{width=\linewidth, center}
        \begin{tabular}{p{0.38\linewidth}ccc}
        \hline\noalign{\smallskip}
                                                            & \textbf{image denoising} & \textbf{image deraining} & \textbf{image dehazing} \\ \noalign{\smallskip}\hline
        \textbf{level of abstraction \newline (intuitive)}           & low                      & mid                      & high                    \\ \hline
        \textbf{num. of down-sampling layers required} & 2-3                      & 3-4                      & 4-5                     \\ \hline
        \end{tabular}
    \end{adjustbox}
    \captionof{table}{Minimum numbers of down-sampling layers required for learning / simulating the visual degradations of different image restoration tasks. Here we demonstrates the saturated numbers of down-and-up-sampling layers using U-Nets, where more than this number the performances do not increase. See Appendix G for more details}
    \label{tab:saturated_layers_on_different_tasks}
\end{figure}

\begin{table*}[t]
    \renewcommand{\arraystretch}{1.5}
    \captionsetup{skip=5pt}
    \centering
        \begin{adjustbox}{width=\textwidth, center}
            \csvautobooktabular[table head=\hline\noalign{\smallskip} & \multicolumn{2}{c}{\makecell{\textbf{SSID-Small-sRGB}\\image denoising}} &\multicolumn{2}{c}{\makecell{\textbf{RESIDE}\\image dehazing}} &\multicolumn{2}{c}{\makecell{\textbf{Rain800}\\image deraining}} & \multicolumn{2}{c}{\makecell{\textbf{Rain1200}\\image deraining}} & \multicolumn{2}{c}{\makecell{\textbf{RainCityScapes}\\deraining + dehazing}} & \multicolumn{2}{c}{\makecell{\textbf{OutdoorRain-8-2}\\deraining + dehazing}} \\\noalign{\smallskip}\hline\csvlinetotablerow\\\hline, table foot=\\\hline]{data/datasets_results_full.csv}
        \end{adjustbox}
        \caption{Evaluation results on image restoration tasks}
        \label{table:experiment_results}
        \vspace{-1em}
\end{table*}

\subsubsection{\textbf{Empirical Evidence of Problem 2}}
\label{subsec:evidence_2}

%\iffalse

\begin{figure}[!htbp]
    \centering
    \begin{subfigure}{.32\linewidth}
        \centering
        \includegraphics[width=.95\linewidth]{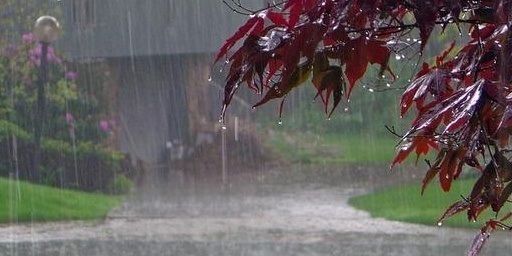}
    \end{subfigure}
    \begin{subfigure}{.32\linewidth}
        \centering
        \includegraphics[width=.95\linewidth]{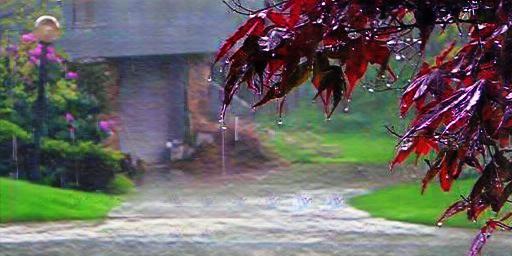}
    \end{subfigure}
    \begin{subfigure}{.32\linewidth}
        \centering
        \includegraphics[width=.95\linewidth]{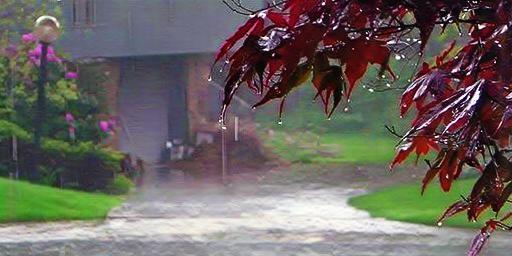}
    \end{subfigure}
    
    \begin{subfigure}{.32\linewidth}
        \centering
        \includegraphics[width=.95\linewidth]{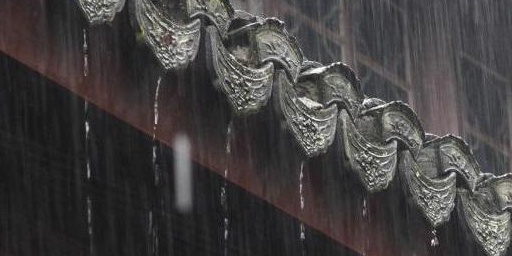}
        \caption{\scriptsize input}
    \end{subfigure}
    \begin{subfigure}{.32\linewidth}
        \centering
        \includegraphics[width=.95\linewidth]{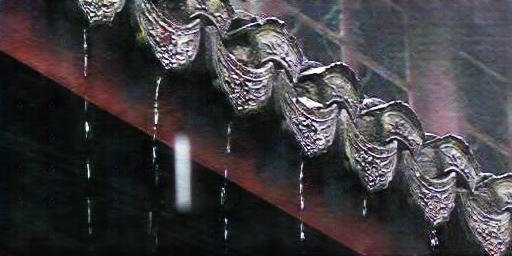}
        \caption{\scriptsize pixel2pixel}
    \end{subfigure}
    \begin{subfigure}{.32\linewidth}
        \centering
        \includegraphics[width=.95\linewidth]{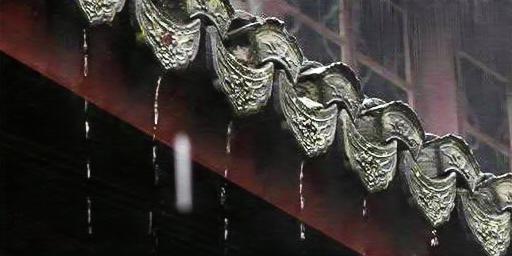}
        \caption{\scriptsize pix2pix + InfoAccum}
    \end{subfigure}
    \captionof{figure}{Examples of image restoration results from conventional DGMs with and without details enhancement on real-world data. Here we demonstrate the examples of deraining results using pixel2pixel \cite{pixel2pixel} model with and without the \textit{InfoAccum} module. We can observe that: even in the areas with no rain, the restored results from these conventional DGMs tend to display inaccurately compared to their corresponding ground truths}
    \label{fig:real_examples}

    \vspace{0.1in}

    \renewcommand{\arraystretch}{1.0}
    \scriptsize
    \begin{adjustbox}{width=\linewidth, center}
            \begin{filecontents*}{data/autoencoder_reconstruct.csv}
,,PSNR,SSIM
En/Decoder,,24.06,0.7194
pix2pix (UNet-8),,34.20,0.9802
UNet-5,,34.28,0.9811
UNet-5 + InfoAccum-15,,42.58,0.9967
UNet-5 + InfoAccum-15 + SubPixUpsamp,,49.41,0.9987
\end{filecontents*}
\csvautobooktabular[table head=\hline\noalign{\smallskip}\csvlinetotablerow\\\noalign{\smallskip}\hline, table foot=\\\hline]{data/autoencoder_reconstruct.csv}
    \end{adjustbox}
    \captionof{table}{Image reconstruction performances of conventional DGMs with and without details enhancement}
    \label{fig:reconstruction_results}
\end{figure}

Although the problem of details loss can be easily observed in their corresponding generated results (Fig. \ref{fig:real_examples}), we also proved it quantitatively by training these generator models directly as AEs (learn to do reconstruction on the input images) (See Table. \ref{fig:reconstruction_results}).

\begin{figure}[!htbp]
    \centering
    \captionsetup{skip=5pt}
        \begin{tikzpicture}
                \begin{axis}
                    [
                    width=0.80\linewidth,
                    height=2in,
                    xtick={0,...,11},
                    ymin=0.75, ymax=0.85,
                    xlabel={\scriptsize number of layers in \textit{InfoAccum} module},
                    ylabel={\scriptsize de-raining performance (SSIM)},
                    xlabel near ticks,
                    ylabel near ticks,
                    xlabel style = {inner sep=0pt},
                    ylabel style = {inner sep=0pt},
                    ticklabel style = {font=\scriptsize},
                    legend style={at={(0.02,0.02)}, anchor=south west, font=\scriptsize}
                    ]
                    \addplot[only marks, color=blue] table [x=num_layers, y=ssim, col sep=comma] {data/num_ddb_layers.csv};
                    \addplot[dashed, domain=-1:11,] {0.0208*ln(x+1) + 0.7817};
                \end{axis}
        \end{tikzpicture}
        \caption{Image restoration performances of baseline DGMs using \textit{InfoAccum} modules with different numbers of layers}
        \label{fig:ddb_layers}
\end{figure}
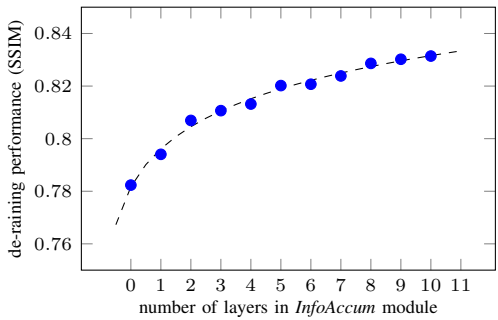

To verify that this problem relates to the loss of low-level information before the decoder, we apply the \textit{InfoAccum} module with different layers to the baseline models and investigate their image restoration performances. Corresponding results (Fig. \ref{fig:ddb_layers}) indicate that the overall performances of models do improve along with the increase of layers in the \textit{InfoAccum} modules. In addition, we also tried to put the \textit{InfoAccum} module in different positions of the DGMs (not only before the encoders) and replaced the \textit{InfoAccum} module with other more complicated network modules (see Appendix G). Relevant results validate that the \textit{InfoAccum} module works as accumulating the inputs' information which does help with the restoration tasks.

To demonstrate that the details loss also happens in the decoder networks, we proposed to enhance the baselines' decoders by adopting the sub-pixel convolution \cite{Sub-pixel_Conv} as the upsampling methods (denoted as \textit{SubPixUpsamp}) of their top layers. Apparent improvements can be observed in the both image reconstruction (Table. \ref{fig:reconstruction_results}) as well as image restoration tasks (Table. \ref{table:experiment_results}).

\subsubsection{\textbf{Empirical Evidence of Problem 3}}
\label{subsec:evidence_3}

%The problem of vanishing gradient and imbalance in GANs can be observed from the training process of these conventional DGMs on image restoration tasks: in all trials of our experiments, the discriminators converged much earlier than the generators, making the training hard to continue. All these training ended up with relatively large values of the generator losses, while the discriminator losses all tend to converge to zero. This is commonly regarded as a failure of training in GANs, where the discriminator fails to provide gradients for the generator to continue training.

%We also observed a two-stage convergence in most of our experiments (Fig. \ref{fig:two_stage_loss}), where the loss functions during the training tend to converge fast in the first stage but suddenly slow down in the second stage. This seems to coincide with our earlier analysis, where the loss function tends to optimize two component objectives with different gradients. We further noticed that pre-training DGMs on image reconstruction tasks before training on image restoration tasks tend to reduce this kind of problem and allow easier convergence.

\begin{figure}[!htbp]
        \centering
        \captionsetup{skip=5pt}
        %\begin{adjustbox}{center}
            \begin{tikzpicture}
                \begin{axis}
                    [
                    width=1.05\linewidth,
                    height=2.5in,
                    xtick={0,25,...,200},
                    xmin=-20, xmax=220,
                    ymin=7.5, ymax=18.5,
                    xlabel={\scriptsize epoch},
                    ylabel={\scriptsize training loss},
                    xlabel near ticks,
                    ylabel near ticks,
                    xlabel style = {inner sep=0pt},
                    ylabel style = {inner sep=0pt},
                    ticklabel style = {font=\scriptsize},
                    legend style={at={(1,1)},anchor=north east, font=\scriptsize, cells={align=left}},
                    legend cell align=left
                    ]
                    \addplot[only marks, color=blue, mark size=1.2pt] table [x=epoch, y=loss, col sep=comma] {data/loss_along_training.csv};
                    \addlegendentry{recorded average loss values};
                    %\addplot[color=blue] table [x=epoch, y=loss, col sep=comma] {data/loss_along_training.csv};
                    \addplot[dashed, domain=-2:225, samples=1000, style=very thick, color=blue] {3.8823*exp(-0.8219*x) + 4.1864*exp(-0.0157*x) + 8.7586};
                    \addlegendentry{fitting curve: $3.9\mathrm{e}^{-0.822x} + 4.2\mathrm{e}^{-0.016x} + 8.8$};
                    \addplot[dashed, domain=-2:225, samples=1000,  style=very thick, color=green] {3.8823*exp(-0.8219*x) + 8.7586};
                    \addlegendentry{1st component: $3.9\mathrm{e}^{-0.822x} + 8.8$};
                    \addplot[dashed, domain=-15:225, samples=1000,  style=very thick, color=red] {4.1864*exp(-0.0157*x) + 8.7586};
                    \addlegendentry{2nd component: $4.2\mathrm{e}^{-0.016x} + 8.8$};
                \end{axis}
            \end{tikzpicture}
        %\end{adjustbox}
        \caption{Two-stage variation of the loss function along the training process with multinomial exponential regression results. More details is explained in Appendix F.}
        \label{fig:two_stage_loss}
        \vspace{-1em}
\end{figure}
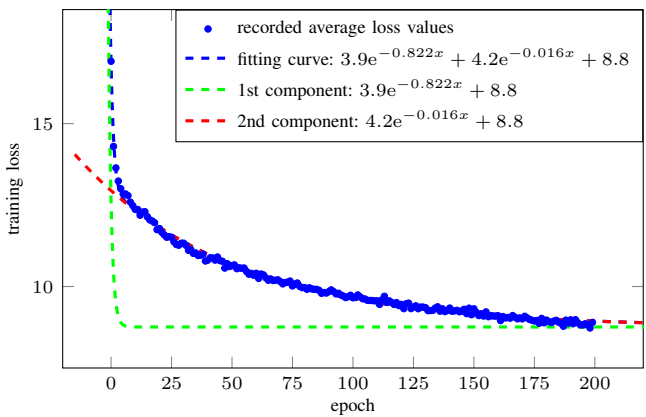

The problem of vanishing gradient and imbalance in GANs can be apparent in the training of these conventional DGMs on image restoration tasks: we observed a two-stage convergence in most of our experiments (Fig. \ref{fig:two_stage_loss}), where the loss functions tend to converge fast in the first stage but suddenly slow down in the second stage. This seems to coincide with our earlier analysis, where the loss functions tend to optimize two component objectives with different gradients. We further noticed that pre-training DGMs on image reconstruction before training on specific restoration tasks can alleviate this kind of problem and allow easier convergence.

For GAN-based methods, all trials of our experiments ended up with large values in the generator losses, while the discriminator losses all tend to approximate zero. This is commonly regarded as a training failure in GANs, where the discriminators converge much earlier than the generators, thus cannot provide gradients for the generators to continue training. To further validate this imbalance problem, we applied LSGAN loss \cite{LSGAN} to replace the traditional GAN loss based on JS-divergence \cite{GAN}, which works as a more sensitive measure when the distributions between the targets the generated results are fairly close to each other. We find that it also allows further convergence of GAN models and significantly improves the restoration performances based on the baselines (Table. \ref{table:experiment_results}).

\subsubsection{\textbf{General Experiments on Image Restoration Tasks}}
\label{subsec:general_experiments}
We generally validate the above problems and the proposed solutions on the benchmarks of different image restoration tasks. Since most existing generative methods tend to base on the U-Net structure, here we applied the pixel2pixel \cite{pixel2pixel} model (\textit{pix2pix}), which uses an 8-layer U-Net, as our baseline. We reduced its over-invested abstraction process by using a 5-layer U-Net (\textit{UNet-5}) (which we found sufficient for most image restoration tasks), equipped it with a 15-layer \textit{InfoAccum} module (\textit{InfoAccum-15}) and modified its decoder with \textit{SubPixUpsamp} to reduce the inherent details loss. Ultimately, we adopted the \textit{LSGAN} loss in replace of the original loss function to validate the vanishing gradient and imbalance training problem. We trained and evaluated the above models on benchmarking datasets of \textit{SIDD-Small} \cite{SIDD_Denoise} for image denoising, \textit{RESIDE-ITS} \cite{RESIDE_Dehaze} for image dehazing, \textit{Rain800} \cite{ID-CGAN_Derain} and \textit{Rain1200} \cite{DID-MDN_Derain} for image deraining, as well as \textit{RainCityScapes} \cite{DAFNet_Derain} and \textit{OutdoorRain-8-2} \cite{HeavyRainRestorer_Derain} for the hybrid of deraining and dehazing. More details about datasets, implementation, and further discussion can be found in the Appendix H. Results (Table. \ref{table:experiment_results}) indicates the proposed solutions achieve apparent improvements with \textit{InfoAccum-15}, \textit{SubPixUpsamp}, and \textit{LSGAN}, with no performance drop on \textit{UNet-5}.

\section{Conclusion}
In this study, we identified three sources of information that are optimized in the generative methods for image restoration and we re-interpret their learning mechanism using information theory. We further pointed out three key issues in the existing methods, gave general solutions, and validated them on the benchmarks of different image restoration tasks.

%In this study, we interpret the learning mechanism of generative methods for image restoration tasks using information theory. We analyzed the information flow in these models and consider there are three sources of information involved in generating the restoration results. Furthermore, we identified three key issues in the direct applications of conventional generative models to image restoration tasks, including over-invested abstraction, inherent details loss, as well as gradient vanishing, and imbalance of training. We pointed out general solutions to these issues and validated their improvement of restoration performance on the benchmarks of different image restoration tasks.

\appendices

\section{Generative Methods and Deconstructive Methods for Image Restoration: Definition \& Comparisons}
In this study, we define generative methods for image restoration as methods that use (conditional) Deep Generative Models (DGMs) \cite{DGM1,DGM2,DGM3} (like Autoencoders (AEs) \cite{AutoEncoder,SparseAutoEncoder,VAE} and Generative Adversarial Networks (GANs) \cite{GAN,DCGAN,CGAN,pixel2pixel,cycleGAN}) or similar deep neural network models that conduct end-to-end simulations of the entire processes of image restoration task(s) as high-dimensional probability distributions on a latent feature space, and generate restoration results by sampling from the distributions conditioned on the visually-degraded inputs. Differently, deconstructive methods are methods that try to only simulate the visual degradations / distortions in the specific image restoration tasks as independent layer(s) of pixels (using either deep-learning-based models or other conventional models) and try to describe their integrations with the background scenes images using handcrafted hypothetical composition models (such as linear additive models et al.). Therefore, whether a handcrafted composition model is involved can be one of the key identities to distinguish between a generative method and a deconstructive method for image restoration.

Compared with deconstructive methods, generative methods try to directly optimize the generated results to approximate the targets' distribution, where both the patterns of visual degradations as well as their integrations with the background scenes are learned as a whole inside the DGMs. Therefore, we do not need an explicit understanding of the properties of the compositions or the detailed mechanisms behind them, and thus handcrafting composition models is no longer required and can elegantly avoid human bias.

Moreover, generative methods also have better support in completing missing details / damaged information from the source inputs. Traditional image restoration methods tend to consider all information for the restored targets can be fully retrieved from their source inputs, where they tend to ignore the details and information that are completely damaged, seriously distorted, or fully covered by the visual degradations and are unrecoverable from the single input data. Therefore, deconstructive methods may not be able to recover / complete the missing information, unless specifically designed by introducing extra networks \cite{DRD-Net_Derain,GraNet_Derain}. Whereas for generative methods, this is functionally well-supported: by transferring general knowledge learned from bid data, DGMs, especially GAN-based models, can easily fill up the missing pixels / lost information and can generate semantically plausible restoration results.

Generative methods also have better generalization ability. Unlike deconstructive methods that tend to be task-specific and require the specialized design of composition models for different tasks, models in generative methods may be generally applicable to different image restoration tasks and even allow ``all-in-one" models. 

In addition, generative methods also benefit from lighter-weight scales and more concise models compared with those deconstructive methods using sophisticated architectures that are based on handcrafted composition models.

To sum up, the advantages of generative methods compared with deconstructive methods, as well as their existing problems are as follows:

%In a nutshell, table \ref{} generally summarizes the advantages of generative methods compared with deconstructive methods, as well as their existing problems.

\textbf{Pros}:
\begin{enumerate}
    \item more accurately simulate the real-world scenarios with sufficient training data and can avoid human bias;
    \item allow learning knowledge from big data to complete damaged details / missing information, and can generate more semantically plausible restoration results with high fidelity;
    \item can be generally applicable to different image restoration tasks without task-specialized designs of composition models, and even allow all-in-one models;
    \item have more concise end-to-end models and lighter-weight in scale without complicated composition models / deconstruction process, which are less likely to overfit;
    \item allow end-to-end training without multi-path / multi-stage optimization, which can have direct gradients and efficient updating of parameters during training and often have much faster inference speed;
    \item GAN-based generative methods can have better support for unsupervised training of models with unpaired or real-world data.
\end{enumerate}
\textbf{Cons}:
\begin{enumerate}
    \item rely on much larger amount of training to converge or achieve competitive performances;
    \item tend to be black-boxes and less interpret-able, thus can be difficult to design network structure or make improvement.
\end{enumerate}

\section{Information Flow \& Training Objectives}
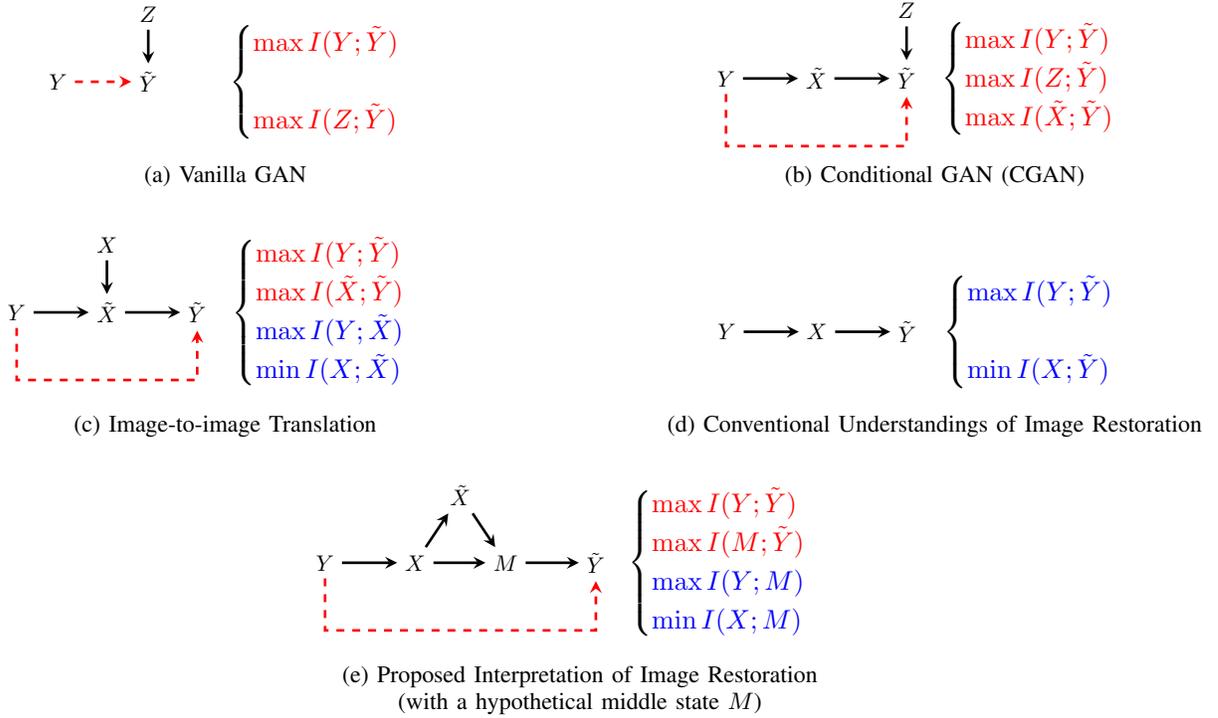
\begin{figure*}[!htbp]
    \centering
    \begin{subfigure}[t]{0.48\textwidth}
        \centering
        \begin{tikzpicture}[scale=0.6, every node/.style={scale=0.85}]
            \centering
            \node at (-3, 0) {};
            \node (Y) at (-2, 0) {$Y$};
            \node (Y_tilde) at (0, 0) {$\tilde{Y}$};
            \node (Z) at (0, 1.5) {$Z$};
            \draw[dashed, -stealth, red, line width=1] (Y) -- (Y_tilde);
            \draw[-stealth, line width=1] (Z) -- (Y_tilde);
            \node at (4, 0) {
                \scalebox{1.2}{
                $\displaystyle
                    \begin{cases}
                      {\color{red}\max I(Y; \tilde{Y})} \\
                      \\
                      {\color{red}\max I(Z; \tilde{Y})} 
                    \end{cases}
                $
                }
            };
        \end{tikzpicture}
        \caption{Vanilla GAN}
        \label{fig:information_bottleneck_generative_models_a}
        \vspace*{5mm}
    \end{subfigure}
    \hfill
    \begin{subfigure}[t]{0.48\textwidth}
        \centering
        \begin{tikzpicture}[scale=0.6, every node/.style={scale=0.85}]
            \node (Y) at (-2, 0) {$Y$};
            \node (X_tilde) at (0, 0) {$\tilde{X}$};
            \node (Y_tilde) at (2, 0) {$\tilde{Y}$};
            \node (Z) at (2, 1.5) {$Z$};
            \draw[dashed, -stealth, red, line width=1] (Y) -- (-2, -1.5) -- (2, -1.5) -- (Y_tilde);
            \draw[-stealth, line width=1] (Y) -- (X_tilde);
            \draw[-stealth, line width=1] (X_tilde) -- (Y_tilde);
            \draw[-stealth, line width=1] (Z) -- (Y_tilde);
            \node at (5, 0) {
                \scalebox{1.2}{
                $\displaystyle
                    \begin{cases}
                      {\color{red}\max I(Y; \tilde{Y})} \\
                      {\color{red}\max I(Z; \tilde{Y})} \\
                      {\color{red}\max I(\tilde{X}; \tilde{Y})}
                    \end{cases}
                $
                }
            };
        \end{tikzpicture}
        \caption{Conditional GAN (CGAN)}
        \label{fig:information_bottleneck_generative_models_b}
        \vspace*{5mm}
    \end{subfigure}
    \hfill
    \begin{subfigure}[t]{0.48\textwidth}
        \centering
        \begin{tikzpicture}[scale=0.6, every node/.style={scale=0.85}]
            \node (Y) at (-2, 0) {$Y$};
            \node (X_tilde) at (0, 0) {$\tilde{X}$};
            \node (Y_tilde) at (2, 0) {$\tilde{Y}$};
            \node (X) at (0, 1.5) {$X$};
            \draw[dashed, -stealth, red, line width=1] (Y) -- (-2, -1.5) -- (2, -1.5) -- (Y_tilde);
            \draw[-stealth, line width=1] (Y) -- (X_tilde);
            \draw[-stealth, line width=1] (X_tilde) -- (Y_tilde);
            \draw[-stealth, line width=1] (X) -- (X_tilde);
            \node at (5, 0) {
                \scalebox{1.2}{
                $\displaystyle
                    \begin{cases}
                      {\color{red}\max I(Y; \tilde{Y})} \\
                      {\color{red}\max I(\tilde{X}; \tilde{Y})} \\
                      {\color{blue}\max I(Y; \tilde{X})} \\
                      {\color{blue}\min I(X; \tilde{X})}
                    \end{cases}
                $
                }
            };
        \end{tikzpicture}
        \caption{Image-to-image Translation}
        \label{fig:information_bottleneck_generative_models_c}
        \vspace*{5mm}
    \end{subfigure}
    \hfill
    \begin{subfigure}[t]{0.48\textwidth}
        \centering
        \begin{tikzpicture}[scale=0.6, every node/.style={scale=0.85}]
            \node (Y) at (-2, 0) {$Y$};
            \node (X) at (0, 0) {$X$};
            \node (Y_tilde) at (2, 0) {$\tilde{Y}$};
            \draw[-stealth, line width=1] (Y) -- (X);
            \draw[-stealth, line width=1] (X) -- (Y_tilde);
            \node at (5, 0) {
                \scalebox{1.2}{
                $\displaystyle
                    \begin{cases}
                      {\color{blue}\max I(Y; \tilde{Y})} \\
                      \\
                      {\color{blue}\min I(X; \tilde{Y})} 
                    \end{cases}
                $
                }
            };
        \end{tikzpicture}
        \caption{Conventional Understandings of Image Restoration}
        \label{fig:information_bottleneck_generative_models_d}
        \vspace*{5mm}
    \end{subfigure}
    \hfill
    \begin{subfigure}[t]{0.95\textwidth}
        \centering
        \begin{tikzpicture}[scale=0.6, every node/.style={scale=0.85}]
            \node (Y) at (-3, 0) {$Y$};
            \node (X) at (-1, 0) {$X$};
            \node (X_tilde) at (0, 1.5) {$\tilde{X}$};
            \node (M) at (1, 0) {$M$};
            \node (Y_tilde) at (3, 0) {$\tilde{Y}$};
            \draw[dashed, -stealth, red, line width=1] (Y) -- (-3, -1.5) -- (3, -1.5) -- (Y_tilde);
            \draw[-stealth, line width=1] (Y) -- (X);
            \draw[-stealth, line width=1] (X) -- (X_tilde);
            \draw[-stealth, line width=1] (X_tilde) -- (M);
            \draw[-stealth, line width=1] (X) -- (M);
            \draw[-stealth, line width=1] (M) -- (Y_tilde);
            \node at (6, 0) {
                \scalebox{1.2}{
                $\displaystyle
                    \begin{cases}
                      {\color{red}\max I(Y; \tilde{Y})} \\
                      {\color{red}\max I(M; \tilde{Y})} \\
                      {\color{blue}\max I(Y; M)} \\
                      {\color{blue}\min I(X; M)}
                    \end{cases}
                $
                }
            };
        \end{tikzpicture}
        \caption{Proposed Interpretation of Image Restoration \\ (with a hypothetical middle state $M$)}
        \label{fig:information_bottleneck_generative_models_e}
        \vspace*{2mm}
    \end{subfigure}
    
    \caption{Information flows and optimization Objectives among Different Generative Models and Tasks}
    \label{fig:information_bottleneck_generative_models}
\end{figure*}

The proposed information-theoretic framework (information flow and its training objectives) for generative methods in image restoration tasks can be inferred by analogy from the information analyses of other conventional DGMs. The information bottleneck principle \cite{Information_Bottleneck,Information_Bottleneck_Deep_Learning_1,Information_Bottleneck_Deep_Learning_2} originally focuses on the information extraction process for discriminative deep neural networks (such as classification, prediction, and dimensional compression), while previous works \cite{Information_Bottleneck_VAE,InfoGAN,IB-GAN} try to generalize it to explain the training process of DGMs. Here we re-analyze and indicate the information flows as well as their optimization objectives of different conventional models and therefore derive our proposed interpretation (Fig. \ref{fig:information_bottleneck_generative_models} compares the relevant framework of these models).

The original GAN model \cite{GAN} (Vanilla GAN) can be regarded as a decoder network that attempts to reach a balance between its inputs $Z$ and the targets $Y$ in the generated results $\tilde{Y}$ (Fig. \ref{fig:information_bottleneck_generative_models_a}). $Z$ here is random noise, which is responsible for adding variations (mainly low-level details) to the generated results and thus is independent of $Y$.

Conditional GAN (CGAN) \cite{CGAN} and InfoGAN \cite{InfoGAN} take extra inputs of conditions / class labels, which is related to the target $Y$ (thus denoted as $\tilde{X}$). Therefore, information of $Y$ is guiding the generation of $\tilde{Y}$ in two paths: $I(Y; \tilde{X}; \tilde{Y})$ and $I(Y \mid \tilde{X}; \tilde{Y})$ (Fig. \ref{fig:information_bottleneck_generative_models_b}).

Models above only play the role of generation, where inputs to the networks (decoder) are already highly condensed. Whereas for generation tasks like image-to-image translation \cite{pixel2pixel,cycleGAN}, inputs to the models (such as images) are of high-dimensionality and involve considerable irrelevant information. Hence the encoder networks are equipped for extracting features $\tilde{X}$ from these inputs $X$ before passing them to the decoders for the generation process. The overall training objectives of the models, therefore, consist of components for both the encoder network (formula in blue: compressing information and fitting the targets' features) and the decoder network (formula in red: optimizing generation results). Notably, the features $\tilde{X}$ to be extracted are supposed to be information in the inputs $X$ that can help with the generation and simulation of the targets $Y$ (i.e. information shared between $X$ and $Y$: $I(Y; X)$). In conventional image-to-image translation tasks, this maintains the consistency of high-level semantics before and after the translation, where the inputs $X$ and the targets $Y$ probably have no dependency nor relation except for the high-level features $\tilde{X}$ they shared (Fig. \ref{fig:information_bottleneck_generative_models_c}).

This can be a different story for the image restoration tasks. The conventional understandings tend to consider that the inputs $X$ (visually-degraded images) are determined by their corresponding targets $Y$ (images of the clean background scenes), where a Markov Chain above stands (Fig. \ref{fig:information_bottleneck_generative_models_d}). It assumes that information of $Y$ is fully contained in $X$, thus the process of extracting relevant information of $Y$ from $X$ is also the process to obtain the restoration results at the same time: $I(Y; X) = H(Y)$. Therefore, there is no need for a network like a decoder to do generation nor to optimize its generated results based on the limited information it received by approximating specific targets.

Of course, in this study, we pointed out that the conventional understanding above can also be less accurate in interpreting the learning process of generative methods in image restoration. As discussed in the body text, the input observed images $X$ to be restored may not contain all the information about the targets $Y$ ($I(Y; X) < H(Y)$), thus a generation process with a decoder network can be essential to provide extra information for fully restoring $Y$. Moreover, different from the traditional image-to-image translation models, information passed to the decoder network for the generation does not only come from the extracted features $\tilde{X}$ by the encoder network but also directly flows from $X$ without abstraction process.

Therefore, we consider the generative methods for image restoration should be understood by putting together both the conventional interpretation as well as the DGMs for image-to-image translation. Suppose there exist a middle state $M$ between the feature extraction process (which tries to extract all the information from the inputs $X$ and stored in $H(M)$) and the generation process (which tries to complete the information that is absent from $X$ but required for restoring $Y$), the information flow can therefore be written as Figure \ref{} and its optimization objectives can therefore be written as both optimizing the two processes respectively (which will be further derived to remove $M$ in the next section).

Noted that $M$ here is DIFFERENT from the middle state between the encoder and the decoder networks. Because we found that: both the information about the background scenes to be retained as well as the information about the features / patterns of the visual degradations are passed from the encoders to the decoders, and both the removal of these visual degradations as well as the completion of missing information / details happen inside the decoder networks.

\section{Proof of Optimization Boundaries}
In this section, we further derived the interpretation in the last section to obtain our proposed information-theoretic framework in this study as well as its component optimization boundaries simultaneously.

Given the information flow and the optimization objectives of the models as Fig. \ref{fig:information_bottleneck_generative_models_e} the forth objective can be divided into optimizing two components:

\begin{equation}
    \min I(X; M) = \min I(X; \tilde{X}; M) + \min I(X \mid \tilde{X}; M)
\end{equation}

\noindent Because in DGMs, $I(X; \tilde{X}; M)$ are features extracted by the learn-able parameters inside the encoder networks, while $I(X \mid \tilde{X}; M)$ is only information in $X$ that directly passes through the encoders without learning process, we can notice that the possible ranges of the two terms above are different (noted that $\tilde{X}$ and $\tilde{Y}$ are the variables to be optimized):

\begin{align}
    -H(\tilde{X}) \le I(X; \tilde{X}; M) \le H(\tilde{X}) \\
    0 \le I(X \mid \tilde{X}; M) \le H(X)
\end{align}

\noindent Thus, as long as $H(\tilde{X}) \ge H(X \mid Y)$, we can easily solve the min-max problem by considering the last two objectives ($\max I(Y;M)$ and $\min I(X; M)$) together:

\begin{equation}
    \begin{cases}
       \min I(Y; X; \tilde{X}; M) \geqslant -H(X \mid Y) \\
       \max I(X \mid \tilde{X}; M) \leqslant H(X)
\end{cases}
\end{equation}

\noindent where the two paths of information that are passed to $M$ are optimized to approximate the information about the features / patterns of visual degradations $-H(X \mid Y)$ and the intact information of the source inputs $H(X)$, respectively, which, add together to approximate the total amount of information about the targets $Y$ that can be retrieved from the inputs $X$ ($H(X) - H(X \mid Y) = I(Y; X)$).

For the first two objectives ($\max I(Y; \tilde{Y})$ and $I(M; \tilde{Y})$), since they are all doing maximization, we can simply get the objectives of the two paths of information that are passed to $Y$:

\begin{equation}
    \begin{cases}
       \max I(M; \tilde{Y}) \leqslant I(Y; X) \\
       \max I(Y \mid M; \tilde{Y}) \leqslant H(Y \mid X)
\end{cases}
\end{equation}

\noindent similarly, the two paths of information add together to approximate the intact information of the targets $Y$ ($I(Y; X) + H(Y \mid X)$).

On account that $M$ here is just a hypothetical middle state, and there does not exist such a variable in the actual models of generative methods in image restoration, we can easily simplify the above information flow and optimization objectives as well as their optimal information boundaries as follow:

\begin{figure}[!htbp]
        \centering
        \begin{tikzpicture}[scale=0.6, every node/.style={scale=0.8}]
            \node (Y) at (-3, 0) {$Y$};
            \node (X) at (-1, 0) {$X$};
            \node (X_tilde) at (0, 1.5) {$\tilde{X}$};
            \node (Y_tilde) at (1, 0) {$\tilde{Y}$};
            \draw[dashed, -stealth, red, line width=1] (Y) -- (-3, -1) -- (1, -1) -- (Y_tilde);
            \draw[-stealth, line width=1] (Y) -- (X);
            \draw[-stealth, line width=1] (X) -- (X_tilde);
            \draw[-stealth, line width=1] (X_tilde) -- (Y_tilde);
            \draw[-stealth, line width=1] (X) -- (Y_tilde);
            \node at (5, 0) {
                \scalebox{1.2}{
                $\displaystyle
                    \begin{cases}
                      \min I(X; \tilde{X}; \tilde{Y}) \\
                      \max I(X \mid \tilde{X}; \tilde{Y} \\
                      \max I(Y \mid X; \tilde{Y})
                    \end{cases}
                $
                }
            };
        \end{tikzpicture}
   \caption{Information flow and the optimization objectives of the proposed interpretation.}
   \label{fig:information_flow_optimization_objectives_simp}
\end{figure}
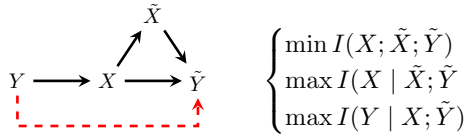

\noindent where the information boundaries of the above optimization objectives are as follow:

\begin{equation}
    \begin{cases}
        I(X; \tilde{X}; \tilde{Y}) \geqslant -H(X \mid Y) \\
        I(X \mid \tilde{X}; \tilde{Y} \leqslant H(X) \\
        I(Y \mid X; \tilde{Y}) \leqslant H(Y \mid X)
\end{cases}
\end{equation}

\section{Explanation of Problem 1: Over-invested Abstraction Process}
Besides the intuitive explanation in the body text, the problem of over-invested abstraction process can also be explain using the theoretic framework.

Suppose the total amount of information required to describe the features / patterns of visual degradations is limited to $H(X \mid Y)$, which is consider to be extracted and passed by the encoder network through $I(X; \tilde{X})$, thus occupying a certain proportion in $H(\tilde{X})$. For CNN-based DGMs, this process of abstraction is often achieved by the down-and-up-sampling mechanism, given the total number of down-and-up-sampling layers in a generator network as $N$ and the corresponding amount of information can be passed through each of these layers as $H(\tilde{X})_{n}^{N}$ $(n \in {1, 2, ..., N})$. For a UNet-like generator network (Encoder-decoder network with skip connections connecting corresponding down-and-up-sampling layers on both sides), the total amount of information can pass through the encoder network: $H(\tilde{X})^{N} = \sum_{n}^{N} H(\tilde{X})_{n}^{N}$, which increase along $N$. When $H(\tilde{X})^{N'} \ge H(X \mid Y)$, continue increasing $N$ may no longer help to extract features of visual degradations for further improving the performance of models on the restoration tasks, causing excessive network parameters, and may even involve extra information of noise $H(\tilde{X} \mid X)$. But for a generator of encoder-decoder without skip-connection, the total amount of information can pass is limited by the bottleneck layers: $H(\tilde{X})^{N} = \min \{H(\tilde{X})_{n}^{N} \mid n \in 1, 2, ..., N\}$, which decrease along $N$. When $H(\tilde{X})^{N'} \le H(X \mid Y)$, continue increasing $N$ will contribute to drops in the model's performances due to less enough information can be passed. Therefore, we deem that for both kinds of generator network, there exist a specific number of down-and-up-sampling $N_{saturated}$ where continuing to increase $N$ may do no good to the overall performance of the model in the image restoration tasks.

\section{Explanation of Problem 2: Inherent Details Loss}
For the problem of inherent details loss, since we regard them as low-level information that is supposed to be retained through the network models without abstraction process, this part of the information is optimized through the objective $\max I(X \mid \tilde{X}; \tilde{Y}) \leqslant H(X) $. Specifically, it is related to both two parts of the models: it is not only determined by the amount of information passed to the decoders but is also restricted by the decoder networks' capability to retain relevant information in the generated results.

We noticed that in real practice, both two steps above involve the loss of low-level information. The information loss inside the decoder can be obvious: as a generative problem, extra information introduced by its network parameters can be inevitable (which is also necessary for approximating the absent information $H(Y \mid X)$): $H(\tilde{Y} \mid X, \tilde{X}) \neq 0$ and $H(Y) = H(X) = H(\tilde{X})$. Thus, only a certain proportion of the information that the decoder receives can be retained in the generated results $H(\tilde{Y})$. Whereas more essentially, considerable low-level information has already been discarded before passing to the decoder. We noticed that the network structure of existing generator models does not support passing intact low-level information to the decoder without occupying $H(\tilde{X})$, even with skip connections: $I(X \mid \tilde{X}; M) \ll H(X \mid \tilde{X})$. Altogether, these two sources of information loss ($H(X \mid M)$ and $H(M \mid \tilde{Y}; X \mid \tilde{X})$) constitute total loss of low-level information in the generated results $\tilde{Y}$ ($H(X \mid \tilde{X}, \tilde{Y})$). Noticeably, in the image restoration tasks, the observed inputs $X$ share large proportions of pixels about the background and relevant details with the target outputs $Y$ ($I(X; Y)$ is much larger than the other generative problems), which, we consider, are mainly low-level information. As a consequence, this inherent discard of low-level information in the generators tends to be more fatal in the restoration task, contributing to a more serious loss of details and distortion of the background scenes in the generated results (lower $I(Y; X \mid \tilde{X}; \tilde{Y})$ thus larger $I(Y; X \mid \tilde{Y})$).

\section{Explanation of Problem 3: Vanishing Gradients \& Imbalanced Training}
Given a generator model that tries to generate data $\tilde{Y}$ based on the inputs to the generator $X$ in a bid to approximate the ideal outputs $Y$, existing measures of loss, both pixel-wise similarities (like MAE or MSE loss) and high-level consistency (like perceptural loss and GAN loss) are trying to optimize the mutual information between $\tilde{Y}$ and $Y$ ($I(Y; \tilde{Y})$). According to the information flow of $\tilde{Y}$, can be divided into two parts of optimization objectives:

\begin{equation}
    \max I(Y; \tilde{Y}) = \max \underbrace{I(Y; X; \tilde{Y})}_{\mathcal{L}_{1}} + \max \underbrace{I(Y; \tilde{Y} \mid X)}_{\mathcal{L}_{2}}
\end{equation}

%\noindent \textit{where expected gradient provided by $\mathcal{L}_{1}$ larger than $\mathcal{L}_{2}$: $\mathbb{E} \nabla \mathcal{L}_{1} > \mathbb{E} \nabla \mathcal{L}_{2}$.}

%\noindent \textit{For conventional generation tasks, inputs $X$ to the generator are often random noise that is independent of the targeted outputs $Y$ or short labels of condition that does not involve much information ($I(X; Y) \approx 0$). Thus, the first part of optimization objectives $\mathcal{L}_{1} = I(Y; X; \tilde{Y})$ is often zero or negligible, and the optimization of the above measures tends to be only maximizing the second term $\mathcal{L}_{2}=I(Y; \tilde{Y} \mid X)$. Nevertheless, for the image-to-image translation in the image deraining task, we consider that the input images $X$ contain far more information about target $Y$ than most of the other generation tasks, which makes $\tilde{Y}$ easy to approximate $Y$ by utilizing this information from $X$, where objective $\mathcal{L}_{1}$ converges much faster than $\mathcal{L}_{2}$ (expected gradient $\mathbb{E} \nabla \mathcal{L}_{1} > \mathbb{E} \nabla \mathcal{L}_{2}$). As a consequence, conventional measures above may result in smaller values or even fail to provide gradients for further improving the generated results (gradient vanishing). For GAN models, these measures of performance may lead to an imbalance between the generator and the discriminator model since the generator may converge more easily by maximizing the information from $X$, while the discriminator needs to be trained with more data and epochs.}

For conventional generation tasks, $\mathcal{L}_{1}$ is often zero or negligible, and the optimization of the above measures tends to be only maximizing $\mathcal{L}_{2}$. Nevertheless, for image-to-image translation in the image restoration task, we consider that the input images $X$ and the target $Y$ shared more information $I(Y; X)$ than most of the other generation tasks, which makes $\tilde{Y}$ easy to approximate $Y$ by utilizing this information from $X$, where objective $\mathcal{L}_{1}$ converges much faster than $\mathcal{L}_{2}$ (expected gradient $\mathbb{E} \nabla \mathcal{L}_{1} > \mathbb{E} \nabla \mathcal{L}_{2}$). As a consequence, conventional measures above may result in smaller values or even fail to provide gradients for further improving the generated results (gradient vanishing). For GAN models, these measures of performance may lead to an imbalance between the generator and the discriminator model.

\section{Details of Methods \& Experiments}
\subsection{Datasets}

In this study, we conducted all training experiments and evaluated relevant models mainly on six benchmarking datasets of image restoration as follows:

\begin{itemize}
    \item one image denoising dataset - \textit{SIDD-sRGB} \cite{SIDD_Denoise}
    \item one image dehazing dataset - \textit{RESIDE} \cite{RESIDE_Dehaze}
    \item two image deraining datasets - \textit{Rain800} \cite{ID-CGAN_Derain} \& \textit{Rain12000} \cite{DID-MDN_Derain}
    \item two datasets with rain and haze appear simultaneously - \textit{RainCityScapes} \cite{DAFNet_Derain} \& \textit{OutdoorRain} \cite{HeavyRainRestorer_Derain}
\end{itemize}

To reduce the computational cost for training, we only use the smallest subset of \textit{SIDD-sRGB} (i.e. \textit{SIDD-Small-sRGB}) for training our models, but we evaluate our models on the entire benchmark of the \textit{SIDD-sRGB} dataset (i.e. \textit{SIDD-Validation-sRGB}). Since the testing set of the \textit{OutdoorRain} dataset is not yet public available, we randomly split its training set with ratio 8:2 (7200:1800) as our \textit{OutdoorRain-8-2} datasets in this paper.

Noticeably, apart from the observed image inputs and their corresponding ground truths, three datasets above: \textit{RESIDE}, \textit{RainCityScapes} and \textit{OutdoorRain} provide additional training data to provide extra supervision for their proposed methods. The \textit{RESIDE} dataset provides the layers of haze for each input, the \textit{RainCityScapes} dataset contains maps of scene depth for each training data, and the \textit{OutdoorRain} provides the ground-truths of rain streak layers, atmosphere light layers, as well as transmittance layers as supervision. All this information is useless for generative models, and we only use the hazy(rainy) inputs and their ground truths for training and evaluation.

Detailed statistics for the datasets are summarized in Table \ref{table:datasets}.

\renewcommand{\arraystretch}{1.5}

\begin{table*}[!htbp]
\scriptsize
\begin{adjustbox}{width=\textwidth, center}

\begin{tabular}{|c|l|l|l|l|}
\hline
\textbf{Dataset} & \multicolumn{1}{c|}{\textbf{Image Restoration Tasks}} & \multicolumn{1}{c|}{\textbf{\# Training Data}} & \multicolumn{1}{c|}{\textbf{\# Testing Data}} & \multicolumn{1}{c|}{\textbf{Composition Method}} \\ \hline
SIDD-sRGB        & Image Denoising                                       & 160 (SIDD-Small)                               & 1,280 (SIDD-Validation)                       & Linear Additive Composition                      \\ \hline
RESIDE   & Image Dehazing                                        & 13,990 (ITS)                                         & 500 (SOTS-indoor)                                         & Atmospheric Scattering Model                     \\ \hline
Rain800          & \multirow{2}{*}{Image Deraining}                      & 700                                            & 100                                           & Linear Additive Composition                      \\ \cline{1-1} \cline{3-5} 
Rain12000        &                                                       & 12,000                                         & 1,200                                         & Density-aware Additive Composition               \\ \hline
RainCityScapes   & \multirow{2}{*}{Image Dehazing + Deraining}     & 9432                                           & 1188                                          & Depth-aware Composition                          \\ \cline{1-1} \cline{3-5} 
OutdoorRain-8-2  &                                                       & 7200                                           & 1800                                          & Heavy Rain Model + Depth-aware Composition       \\ \hline
\end{tabular}

\end{adjustbox}
\caption{Information of the datasets used in our experiments}
\label{table:datasets}
\end{table*}

\subsection{Evaluation Metrics}

We adopted the peak signal to noise ratio (PSNR) and structural similarity index (SSIM) \cite{PSNR_SSIM,ssim} as the quantitative methods to evaluate the performances of models on both image restoration tasks and image reconstruction task. For both PSNR and SSIM, larger values indicate better performances of models.

\subsection{InfoAccum Module}

Rather than directly increasing the information in these skip connections $H(X \mid \tilde{X})$, we proposed to increase the total amount of information in the inputs $H(X)$ as an alternative solution. For the low-level information we intend to enhance, there is:

\begin{equation}
    H(X \mid \tilde{X}) = H(X) - I(X; \tilde{X})
\end{equation}

\noindent Since $I(X; \tilde{X})$ is supposed to be the features of the image degradations, we consider it to be constant. Therefore, we can simply increase the amount of information of inputs $H(X)$ before sending them to the generator networks to indirectly increase $H(X \mid \tilde{X})$ without modifying the skip connections or network structure of the generator network.

More specifically, we proposed to introduce a network module that can enhance the extraction and accumulation of information before sending them to the generator network. We refer to the network structure of Densely Connected Network (DenseNet) \cite{DenseNet}: by using concatenative skip connections, feature maps in the previous layer can be reused in the deeper layers of the network. Thus, source information from the inputs can be fully retained and repeatedly emphasized for further extraction. For a given input $\mathbf{x}_{0}$, the output of a Dense Block can be represented as a recursive concatenation of $L$ layers:

\begin{equation}
    \mathbf{x}_{l}=concat(\left [ \mathbf{x}_{l-1}, F_{l}(\mathbf{x}_{l-1}) \right ])
\end{equation}

\noindent where $F_{l}(\cdot)$ denotes the operations in dense layer $l$.

Notably, by considering the outputs from all $L$ layers as a whole,  $\Psi(\mathbf{x}_{0})$, where $\Psi(\mathbf{x}_{0})=concat(\left [F_{1}(\mathbf{x}_{0}), F_{2}(\mathbf{x}_{1}), \dots, F_{l}(\mathbf{x}_{l-1}) \right ])$ represents a concatenation of extracted feature maps from each layer, the entire outputs of this kind of structure can be regarded as a concatenation of the input $\mathbf{x}_{0}$ and these extracted features $F_{l}(\mathbf{x}_{l-1})$ from each layer: $\mathbf{x}_{l}=concat(\left [ \mathbf{x}_{0}, \Psi(\mathbf{x}_{0})) \right ])$. It indicates that the original input $\mathbf{x}_{0}$ is preserved in its entirety through a direct connection from the beginning to the end, where the later processes can still have intact information of the original source input.

The Residual Network (ResNet) \cite{ResNet} also has a similar network structure by using skip connections to pass information to deeper layers. However, it achieves in an additive manner, which applies in-place addition of the learned residual features with the layer's input. Therefore the output feature maps may hardly contain intact input information for later processing.

\begin{figure}[!htbp]
    \begin{subfigure}{1.0\linewidth}
        \centering
        \includegraphics[width=.95\linewidth]{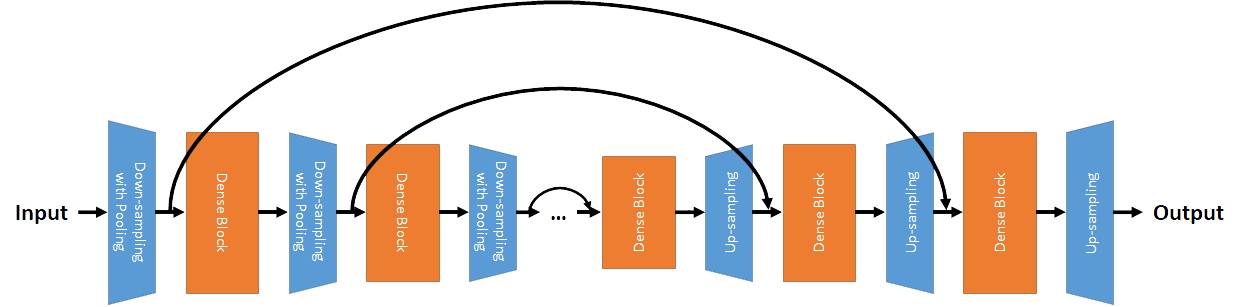}
        \caption{}
    \end{subfigure}
    \begin{subfigure}{1.0\linewidth}
        \centering
        \includegraphics[width=.95\linewidth]{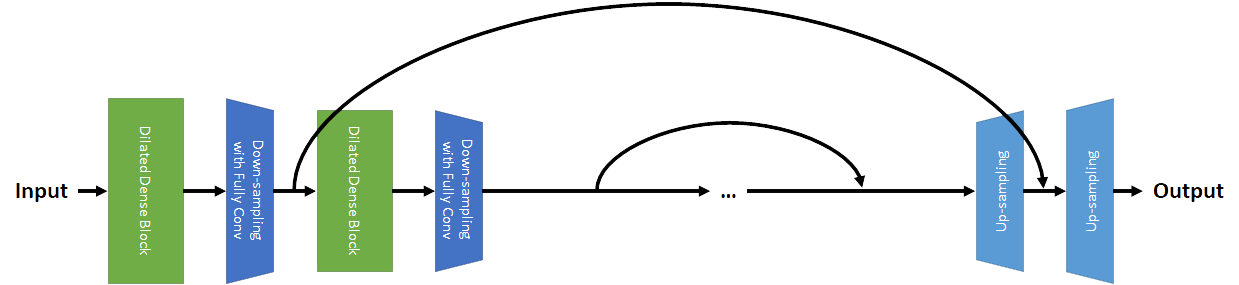}
        \caption{}
    \end{subfigure}
    \caption[Comparison between the generator network of ID-CGAN with DenseNet structure and our proposed detail-enhancing generator.]{Comparison between the generator network of ID-CGAN with DenseNet structure and our proposed detail-enhancing generator using InfoAccum module(s). }
    \label{fig:network_compare}
\end{figure}

Many existing methods also adopted the DenseNet structure in their models, but here we use it differently. For example, Zhang et al. \cite{ID-CGAN_Derain} also adopted the DenseNet structure in their GAN-based deraining model. However, instead of placing the DenseNet module before the down-sampling processes to emphasize the input information, it applies dense blocks after the pooling layers of the network, where details information might have already been lost in the foregoing down-sampling process. Figure \ref{fig:network_compare} illustrates the difference between the previous model and our idea.

Furthermore, we consider some fine-grind details within the patches of the convolutional filters may be obfuscated and hardly recovered if all filters are of the same size. To help with the extraction of these features and to eliminate the interference caused by the difference in receptive fields, we adopt the idea of multi-scaling, so as to aggregate contextual information from different receptive fields. More specifically, we refer to the Dilated Convolution \cite{Dilated_Conv} to obtain a larger receptive field without increasing the number of layers or involving extra parameters and achieve the above idea by using a multi-path structure, concatenating convolutions with different dilation rates.

Our proposed InfoAccum module is indicated as follow:

\begin{equation}
    \mathbf{x}_{l}=concat(\left [ \mathbf{x}_{l-1}, F_{l}(\mathbf{x}_{l-1}), G_{l}(\mathbf{x}_{l-1}), H_{l}(\mathbf{x}_{l-1}) \right ])
\end{equation}

\noindent with $F_{l}(\cdot)$, $G_{l}(\cdot)$ and $H_{l}(\cdot)$ represents the composite functions involving convolution with dilation rate $1$, $3$, and $5$ respectively.

Due to the reuse-ability of features, each dense layer only needs to focus on extracting a small number of features, and the overall feature extract-ability of the module can be determined only by the number of dense layers inside. Theoretically, the number of feature maps in the output is related to both the growth rate and the number of layers. But in this case, a larger growth rate is equivalent to adding extra layers, because the inputs to all the dense layers include complete source data and thus the features extracted from each layer are independent. Therefore, we simply assign a relatively small value to the growth rate and determine the complexity of the feature to be extracted by adjusting only the number of layers, so as to control the feature extract-ability.

\begin{figure}[!htbp]
    \centering
    \includegraphics[width=.95\linewidth]{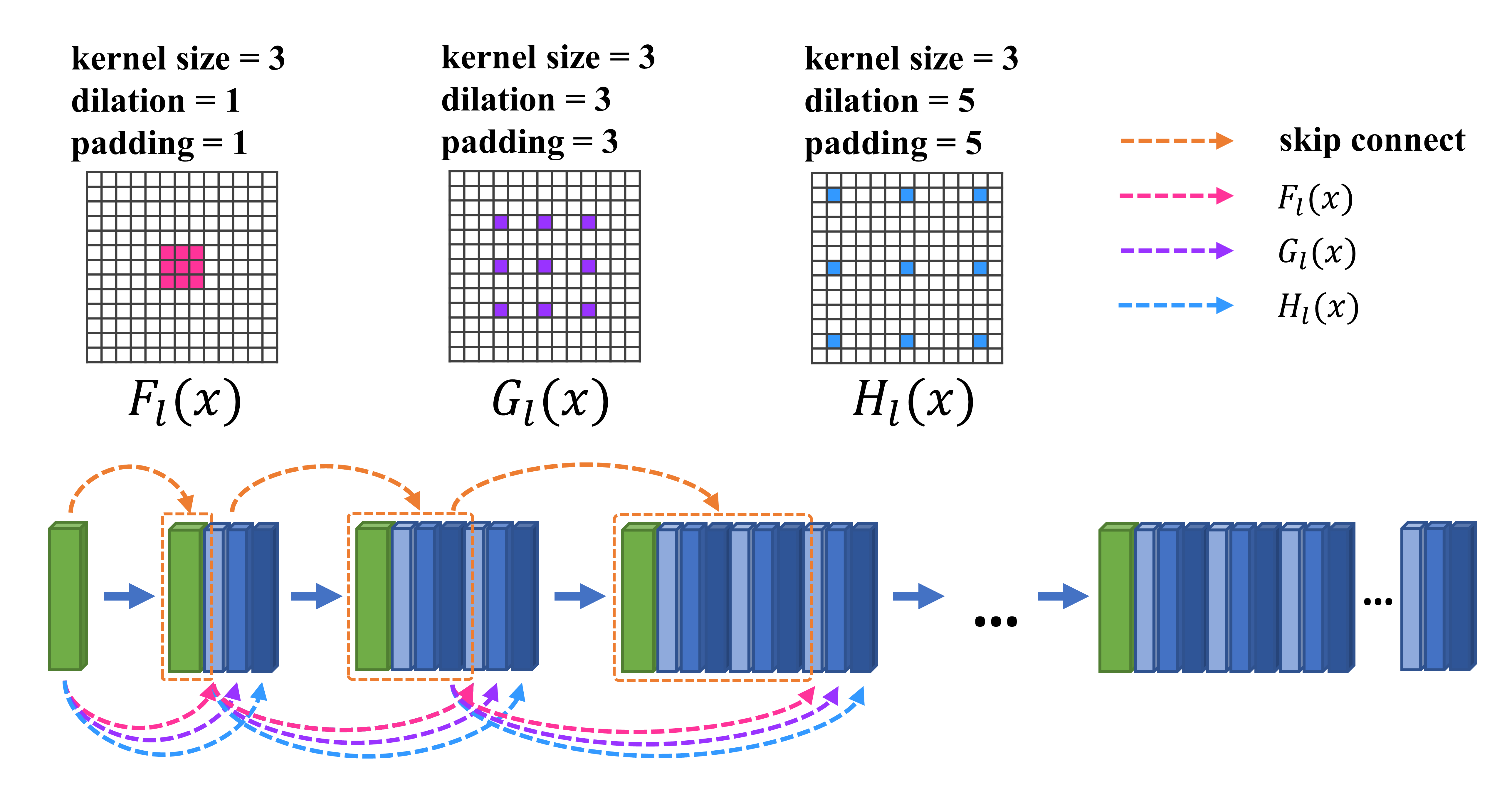}
    \caption{Proposed InfoAccum module}
    \label{fig:ddb}
\end{figure}

\subsection{Sub-pixel Convolutional Upsampling}

The loss of details also exists in the up-sampling process of the decoder network. The earliest up-sampling methods based on un-pooling (missing pixels are abandoned), or interpolation operations (missing pixels are filled based on their neighbors) involve irreversible information loss. Better solutions try to fill the missing pixels with spatially-adjacent textures, or contextual information. For example, in the UNet of the pixel2pixel model, up-sampling is achieved using deconvolution (transposed convolution), which is useful for involving some more general information when filling the missing pixels. However, all these up-sampling methods do not retain the input details and try to fill the missing pixels with calculated results, which is likely to introduce noises or information that is inconsistent with the source inputs, or contributes to the Checkerboard Artifacts \cite{Checkerboard_Artifact} in the generated results.

Sub-pixel convolution \cite{Sub-pixel_Conv} is a better solution for up-sampling, which is commonly used in applications like super resolution for generating higher quality images. A sub-pixel convolution module often consist of a convolution layer and a pixel-shuffle operation, in which an input of $H \times W \times C r^2$ tensor will be rearrange to form a $rH \times rW \times C$ tensor using phase shift ($r$ denotes the upscale factor):

\begin{equation}
    \mathcal{PS}(T)_{h,w,c} = T_{\lfloor h/r \rfloor,\lfloor w/r \rfloor, c \cdot r \cdot mod(w,r) + c \cdot mod(h,r) + c}
\end{equation}

\noindent where h, w and c corresponds to the height, weight and number of channels in the resulted image.

%为了使生成的图像能够最大程度地保留原始输入的细节同时避免生成的图像中出现Checkerboard Artifact，我们直接采用上述的sub-pixel convolution替代传统GAN网络中的deconvolution操作作为模型的上采样方法
To retain details in the generated images to the largest extent and prevent the Checkerboard Artifact, we proposed to use the sub-pixel convolution up-sampling (SubPixUpsamp) at the top layer of the decoder network.

\subsection{General Implementation Details}

Generally in this study, we conduct our experiments mainly based on the pixel2pixel model \cite{pixel2pixel}. Thus, after applying the LSGAN loss, the overall training objectives of the GAN model are as follow:

\begin{equation}
    \mathcal{L}_{\mathcal{G}}(\mathcal{G}, \mathcal{D})=\mathbb{E}_{\mathbf{x}, \mathbf{y}}\left [(\mathcal{D}(\mathbf{x}, \mathcal{G}(\mathbf{x}))-1)^{2} \right ] + \lambda \mathcal{L}_{L1}(\mathcal{G})
    \label{eq:g}
\end{equation}

\begin{equation}
    \mathcal{L}_{\mathcal{D}}(\mathcal{G}, \mathcal{D})=\frac{1}{2}\mathbb{E}_{\mathbf{x}, \mathbf{y}}\left [(\mathcal{D}(\mathbf{x}, \mathbf{y})-1)^{2} \right ]
     + \frac{1}{2}\mathbb{E}_{\mathbf{x}, \mathbf{y}}\left [ \mathcal{D}(\mathbf{x}, \mathcal{G}(\mathbf{x}))^{2} \right ]
\end{equation}

Similarly, we also include L1-loss as the complement to the discriminator on scoring low-frequency information, which can reduce blurring and guide the generator in details adjustment. In case the discriminator fails, the generator can still go in the gradient-appropriate direction.

\begin{equation}
    \mathcal{L}_{L1}(\mathcal{G})=\mathbb{E}_{\mathbf{x}, \mathbf{y}} \left \| \mathbf{y} - \mathcal{G}(\mathbf{x}) \right \| _{1}
\end{equation}

the L1-loss $\mathcal{L}_{L1}(\mathcal{G})$ is joined with the LSGAN MSE loss to form the generator loss (equation \ref{eq:g}), with $\lambda$ as a hyper-parameter.

As for the discriminator, we use a 5-layer fully convolutional network and follow the idea of PatchGAN in pixel2pixel. Since we have mentioned that an imbalance exists between the generator and the discriminator, in which the discriminator is always the first to converge and thus fails to provide the gradient to the generator to continue training. A common understanding here is that the discriminator is over-powerful than the generator. However, we also tried to reduce the number of layers and try to use some "weaker" networks as the discriminator, but all experiment ends up the same. This may illustrate that the difference between the generated data and the real ground truth does not lie on high-level features, and a shallow network can also tell their differences. Therefore, instead of elaborating the discriminator network, we try to reinforce the generator network so as to compete with the discriminator. PatchGAN here is found still useful in deraining tasks, which processes each image patch identically and independently and guarantees that when the noise is not uniformly distributed on the input image, the discriminator can still make a general evaluation on the quality of the generated image. We also compared its performance with the multi-scale discriminator proposed in the ID-CGAN, where the experiment results turn out to be the same. So here for faster training, we do not use the multi-scale model, which includes more convolution operations.

For the training of our model, we use batch size equal to 1. For each iteration of training, image are randomly crop into a smaller size as input to the our model, in a bid to augment the training data and improve model's generalization ability. The ideal crop size are combinations among 256, 512, and 1024, which mainly depends on the datasets (the crop size should be large enough to contain as complete semantic information as possible, the minimum crop size for RainCityScapes dataset, for instance, should be 512x512). we employed Adam as the optimizer with 0.0002 as learning rate, 0.5 and 0.999 as the first and the second momentum values, and 0 as weight decay.

Relevant programmes are implemented using the platform of PyTorch and we conducted all experiments a physical environment with Intel Xeon(R) Silver 4108 as CPU and GeForce RTX 2080 Ti as GPU.

\subsection{Empirical Evidences of Problem 1: Implementation Details}

In these experiments, we adjusted the number of network layers for down-and-up-sampling in the generator models and investigate their corresponding image restoration performances on four datasets above, which are SIDD-sRGB, RESIDE, Rain800, and Rain12000. We adopt two common types of backbone generator networks for comparison: Convolutional Encoder-decoder without skip connection (denoted as \textit{En/Decoder}, which, conventionally, uses $2*2$ max-pooling as the down-sampling method, and nearest-neighbor interpolation for upsampling) and \textit{UNet} \cite{UNet} (which is first introduced as a generator in the pixel2pixel model \cite{pixel2pixel} with 8 layers by default, using fully convolutional layers for both down-sampling and up-sampling with skip connections concatenating outputs of each level). The scale factors of both networks are set to 2, and to be consistent, we compared the performances of both models with 1 to 8 layers of down-and-up-sampling respectively.

Relevant results indicate that: for the \textit{En/Decoder} generators, the restoration performances of models sharply drop after a short climbing (at around 2-3 layers) along with the increase of down-and-up-samplings layers. It may be because the amount of information that can pass is limited by the bottleneck of the network, where information compressed more than 2 layers may not be enough for restoring the clean-background images. This may also reflect that considerable low-level information is required for the restoration task. For the UNet generators, we can observe that the performances of the models increase from 1 to 4 layers, meaning that along with the increase of the number of layers, higher-level features can be extracted, while information from the previous layers can still pass through the skip-connections. Whereas, when reaching around the 5th layer, the performances of models will no longer improve even if we continue to increase the number of layers. It indicates that the level of features to learn to reach their saturation here, and a 5-layer UNet can already achieve the same performance as the 8-layer UNet used in the pixel2pixel models. Extra abstraction processing may not be helpful for the tasks.

The following Figure \cite{} indicate the results of models on three different image restoration tasks.

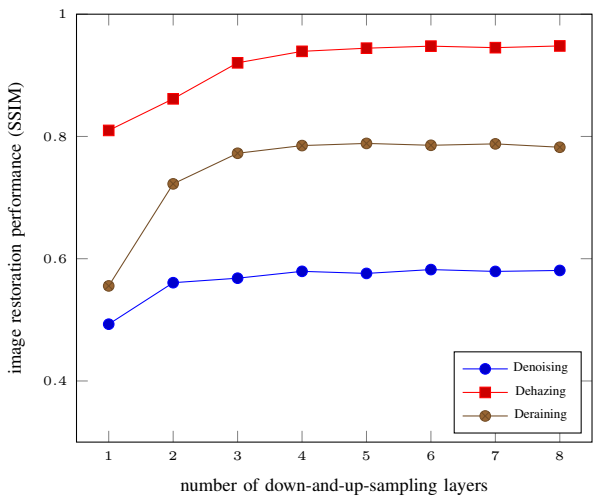
\begin{figure}[!htbp]
        \centering
        \begin{adjustbox}{center}
            \begin{tikzpicture}
                \begin{axis}
                    [
                    xmin=0.5, xmax=8.5, xtick={1,...,8},
                    xlabel={\scriptsize number of down-and-up-sampling layers},
                    ymin=0.3, ymax=1.0,
                    ylabel={\scriptsize image restoration performance (SSIM)},
                    xlabel near ticks,
                    ylabel near ticks,
                    ticklabel style = {font=\tiny},
                    legend style={at={(0.98,0.02)}, anchor=south east, font=\tiny}
                    ]
                    \addplot table [x=num_layers, y=SSIM, col sep=comma] {data/SIDD-Small_unet_test_SSIM.csv};
                    \addplot table [x=num_layers, y=SSIM, col sep=comma] {data/RESIDE-ITS_unet_train_SSIM.csv};
                    \addplot table [x=num_layers, y=SSIM, col sep=comma] {data/Rain800_unet_test_SSIM.csv};
                    \legend{Denoising, Dehazing, Deraining}
                \end{axis}
            \end{tikzpicture}
        \end{adjustbox}
        \caption{Performances of GAN models on different image restoration tasks using U-Net generator networks with different numbers of down-and-up-sampling layers}
        \label{fig:unet_layers_image_restoration_tasks}
\end{figure}

\subsection{Empirical Evidences of Problem 2: Implementation Details}

Three sets of experiments were conducted in this section. We first prove the existence of inherent details loss quantitatively by training relevant generator networks as Autoencoders to perform image reconstruction. We input the clean background images (ground truth) from the Rain800 dataset to these generator networks and have them attempt to output images that are as similar to their inputs as possible by introducing MSE loss between the inputs and outputs. The higher similarity between the inputs and the outputs indicates less information is lost in the generator network. We compare corresponding reconstruction performances of different generator networks. The results show that none of the generator networks can completely reconstruct the input images meaning that all of them more or less suffer from the problem of details loss. Noticeably, the model with details enhancement (methods proposed above) achieves the best restoration performance, with an average PSNR of the restored images reaching 50.0 and an average SSIM reaching 0.9990.

To prove that the details loss problem originates from the discard of low-level information before the decoder network, as well as to verify that improving the extraction and accumulation of information in the inputs can help with the problem, we conducted the second experiment. We adjusted the number of dense layers in the network module equipped before the generator network and investigate their corresponding deraining performances on the Rain800 dataset. Here we use a 5-layer-UNet as the backbone generator and adopt the InfoAccum module to enhance the inputs' information before being sent to the generator network. Relevant results indicate that: as the number of InfoAccum layers increases, the deraining performances of the models also improve significantly. Since the amount of information that actually passed inside the generator networks is constant in all these models, the InfoAccum modules applied only increase the information of the inputs, which, we consider, is acting the role of emphasizing the low-level information in the inputs.

To prove that the details loss also happens inside the decoder network, and enhancing the decoder network can help, we introduced the SubPixUpsamp module to the decoder network and compare its performances on both image deraining and image reconstruction with models without SubPixUpsamp (similar experiment settings as above). Results on decoders with SubPixUpsamp module indicate advances on both deraining and reconstruction performances, meaning that enhancing the decoder network does help to alleviate the problem of details loss.

\section{Discussion \& Supplementary Experiments}
\subsection{Positions of Adding the Detail Enhancing Module}

\begin{figure}[!htbp]
    \centering
    \begin{tikzpicture}
        \begin{axis}
            [
            width=0.5\textwidth,
            height=6cm,
            yshift=-10cm,
            xtick={0,...,8},
            xticklabels={None, 1st, 1st \& 2nd, 1st - 3rd, 1st - 4th, 1st - 5th, 1st - 6th, 1st - 7th, 1st - 8th},
            x tick label style={rotate=45, anchor=east},
            ymin=0.75, ymax=0.85,
            xlabel={\scriptsize position(s) of inserting detail enhancing module},
            ylabel={\scriptsize de-raining performance (SSIM)},
            xlabel near ticks,
            ylabel near ticks,
            ticklabel style = {font=\scriptsize},
            legend style={at={(0.02,0.02)}, anchor=south west, font=\scriptsize}
            ]
            \addplot[only marks, color=red] table [y=ssim, col sep=comma] {data/ddb_insert_slots.csv};
            \addplot[dashed, domain=-1:9,] {-((x+2.99242)^(-3.29221)) + 0.809102};
        \end{axis}
    \end{tikzpicture}
    \caption{Deraining performances of pixel2pixel models \cite{pixel2pixel} with InfoAccum modules inserted to different positions of their backbone generator networks}
    \label{fig:slots_to_insert_ddb}
\end{figure}
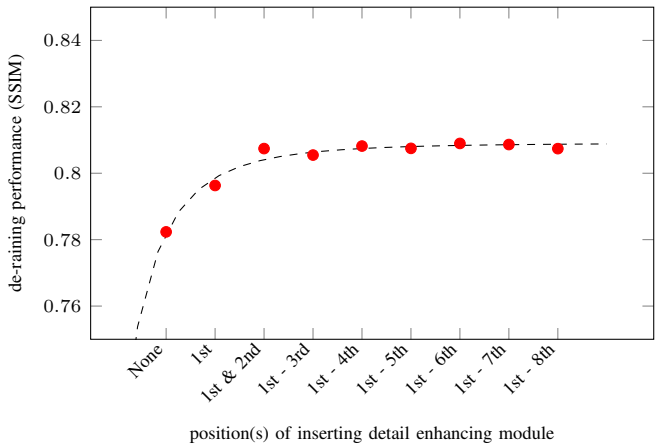

Originally, we intend to apply the InfoAccum module before the generator network to help extraction and accumulation of low-level information. We also investigate the models' performances by applying the network modules on different positions of the generator networks (Fig. \ref{fig:slots_to_insert_ddb}). Here we use the 8-layer-UNet in pixel2pixel \cite{pixel2pixel} as the backbone generator and try to insert 15-layer-InfoAccum modules before each of its encoder layers (``1st" denotes adding a InfoAccum module before the generator, while ``1st - 8th" means that 8 InfoAccum modules are added before all 8 encoder layers of the UNet generator). Similarly, we train the pixel2pixel model on the Rain800 datasets.

We observe that adding the InfoAccum module to the ``1st" position brings the greatest improvement while adding which to deeper layers does not may much different to the restoration performance of the model. This also reveals the inherent discards of low-level information in the network structure before the decoder network. Noticeably, adding extra InfoAccum at the ``2nd" position also make minor improvement on the models. It may indicate that some relatively higher-level information is also enhanced by the InfoAccum module.

\subsection{InfoAccum Modules Compared with other different Network Modules}

\begin{figure}[!htbp]
    \centering
    \begin{tikzpicture}[thick,scale=1, every node/.style={scale=0.8}]
    \begin{axis}
        [
            ybar,
            enlargelimits=0.15,
            ylabel={de-raining performance (SSIM)},
            symbolic x coords={baseline, Residual \cite{DetailNet_Derain_2}, ContextDilated \cite{JORDER_Derain}, SCAN \cite{RESCAN_Derain}, Recursive \cite{LPNet_Derain}, Attention \cite{AttentiveGAN_Derain}, Dense, \textbf{InfoAccum}},
            xtick=data,
            xticklabel style = {rotate=90,anchor=east},
        ]
        \addplot [draw, fill=blue] coordinates {(baseline, 0.7878446588892444) (Residual \cite{DetailNet_Derain_2}, 0.7590201503806622) (ContextDilated \cite{JORDER_Derain}, 0.7943166436427203) (SCAN \cite{RESCAN_Derain}, 0.8122411751956207) (Recursive \cite{LPNet_Derain}, 0.8182004565331423) (Attention \cite{AttentiveGAN_Derain}, 0.8357147741599125) (Dense, 0.8460905507939769) (\textbf{InfoAccum}, 0.84958517409669)};
        \coordinate (baseline_coordinate) at (axis cs:baseline, 0.7878446588892444);
        \coordinate (bottom_left) at (rel axis cs:0,0);
        \coordinate (bottom_right) at (rel axis cs:1,0);

        \draw [red,sharp plot,dashed] (baseline_coordinate -| bottom_left) -- (baseline_coordinate -| bottom_right);

    \end{axis}
    \end{tikzpicture}
    \caption{Deraining performance of models with other de-raining network modules added before the encoder of the baseline generator model.}
    \label{fig:network_modules}
\end{figure}
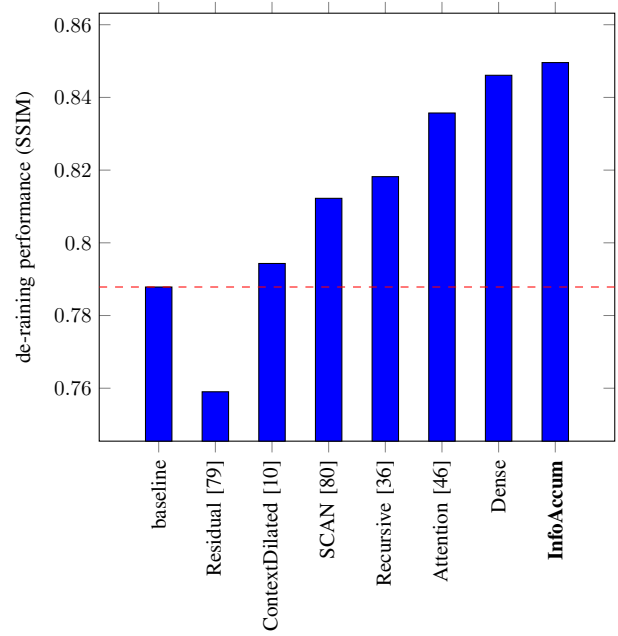

Relevant studies have also proposed considerable network modules to enhance the image restoration performances of their models. Here, we also compare the InfoAccum with some other famous modules proposed for single image deraining task on deraining datasets (Fig. \ref{fig:network_modules}), including Residual deraining module (Residual) \cite{DetailNet_Derain_2}, Contextualized Dilated Block (ContextDilated) \cite{JORDER_Derain}, SCAN module (SCAN) \cite{RESCAN_Derain}, Recursive deraining module (Recursive) \cite{LPNet_Derain}, Attentive Recurrent module (Attention) \cite{AttentiveGAN_Derain}, and ordinary DenseNet module (Dense), as contrast to the proposed InfoAccum module. We use a 5-layer-UNet as backbone generator with SubPixUpsamp module and compare both their image restoration and image reconstruction performances. Results indicate that the proposed InfoAccum module makes the greatest improvement on the baseline model than other network modules.

\vfill

%\bibliography{references.bib}
%\bibliographystyle{IEEEtran}


\end{document}